\documentclass[10pt,twocolumn,letterpaper]{article}

\usepackage{cvpr}
\usepackage{times}
\usepackage{epsfig}
\usepackage{graphicx}
\usepackage{amsmath}
\usepackage{amssymb}
\usepackage{url}
\usepackage{amssymb}
\usepackage{times}
\usepackage{epsfig}
\usepackage{subfigure}
\usepackage{multirow}
\usepackage{algorithmic}
\usepackage{enumerate}
\usepackage{algorithm}
\usepackage{caption}
\newcommand{\R}{{\mathbb R}}
\newcommand{\sphere}{{\mathbb S}}
\DeclareMathOperator*{\argmin}{argmin}

\usepackage[pagebackref=true,breaklinks=true,letterpaper=true,colorlinks,bookmarks=false]{hyperref}

\def\cvprPaperID{1716} 

\cvprfinalcopy
\ifcvprfinal\fi
\begin{document}

\title{Representing Data by a Mixture of Activated Simplices}

\author{\small Chunyu Wang\\
\small Peking University
\and
\small John Flynn\\
\small UCLA
\and
\small Yizhou Wang\\
\small Peking University
\and
\small Alan L. Yuille\\
\small UCLA
}
\maketitle

\begin{abstract}
We present a new model which represents data as a mixture of simplices. Simplices are geometric structures that generalize triangles. We give a simple geometric understanding that allows us to learn a simplicial structure efficiently. Our method requires that the data are unit normalized (and thus lie on the unit sphere). We show that under this restriction, building a model with simplices amounts to constructing a convex hull inside the sphere whose boundary facets is close to the data. We call the boundary facets of the convex hull that are close to  the data Activated Simplices. While the total number of bases used to build the simplices is a parameter of the model, the dimensions of the individual activated simplices are learned from the data. Simplices can have different dimensions, which facilitates modeling of inhomogeneous data sources. The simplicial structure is bounded --- this is appropriate for modeling data with constraints, such as human elbows can not bend more than 180 degrees. The simplices are easy to interpret and extremes within the data can be discovered among the vertices. The method provides good reconstruction and regularization. It supports good nearest neighbor classification and it allows realistic generative models to be constructed. It achieves state-of-the-art results on benchmark datasets, including 3D poses and digits.
\end{abstract}

\section{Introduction}
The curse of dimensionality motivates machine learning researchers to search for low-dimensional structures in high-dimensional data.  These low-dimensional structures help us understand the data, and they enable us to build data models that avoid over-fitting.

There are several strategies for finding low-dimensional structures.  One approach is to project the data into a single low-dimensional linear or non-linear manifold  \cite{tenenbaum2000global, Saul:2003us, Ham:2004ut, Belkin:2003va, Donoho:2003tu} while preserving properties of a local neighborhood graph. These methods are global, in that a single low dimensional structure is used in the representation of the data. One drawback of those methods is that it is difficult to represent some closed manifolds in familiar spaces (such as Euclidean spaces) of a useful low dimension. These closed manifolds are prevalent in computer vision, for example, in data from repeated human walking motion.

A second approach is to represent data using a combination of simple local structures in the original data space. $k$-means and  its variants represent data by assigning data points to one of $k$ clusters. These methods provide coarse piecewise constant approximations to high-dimensional structures. More recently, $k$-flats \cite{Canas:2012vf, Bradley:2000cm}, Atlas learning \cite{Pitelis:2013ik},  and Sparse Subspace Clustering \cite{Elhamifar:2013uz}  represent data using a vocabulary of hyperplanes. These methods work very well when the data come from a union of hyperplanes, but are limited when data are from a curved manifold. Since the hyperplanes are unbounded,  they don't provide for strict constraints, such as might be appropriate for range of motion restrictions in human pose data.

\begin{figure}
\centering
\includegraphics[height=1.2in]{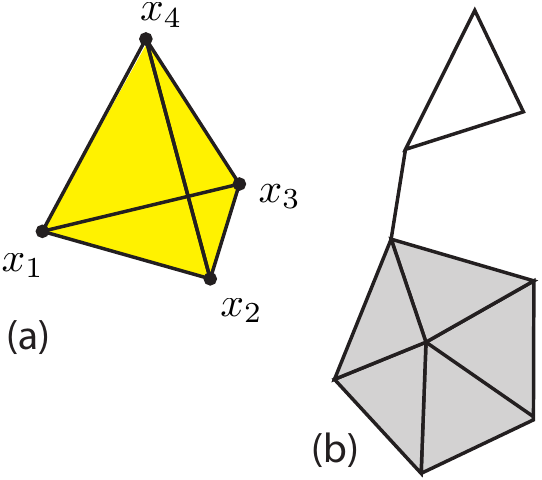}
\caption{(a) A 3-simplex is a solid tetrahedron. Points within the tetrahedron are convex combinations of the 4 vertices $\alpha_1 x_1 + \cdots \alpha_4 x_4$, where $\alpha_i \ge 0$ and $\sum \alpha_i=1$.
(b) A complex consisting of five 2-simplices and some 1-simplices.}
\label{fig:simplex}
\vspace{-1em}
\end{figure}

In this paper we represent data using an arrangement of simplices. Simplices are points, line segments, triangles, tetrahedra, and other higher dimensional analogs of the triangle; see Figure \ref{fig:simplex}.  The simplices act as local models of the data manifold.  In general, learning a model which represents data using a mixture of simplices is quite difficult. However, we show in Section \ref{sec:activated_simplices} that, under some reasonable assumptions, the optimal simplices have a convenient geometry, which makes learning dramatically easier.

We assume that the data are normalized to have norm one and hence lie on a unit sphere, and we build the simplices using bases that have norm at most one. With these restrictions, learning the best simplices amounts to finding boundary facets on the convex hull of the bases that are close to the data. We call the boundary facets activated in this way activated simplices.  Once a collection of activated simplices is learned they are  pruned to prevent over-fitting.


\begin{figure}
\includegraphics[scale =0.65]{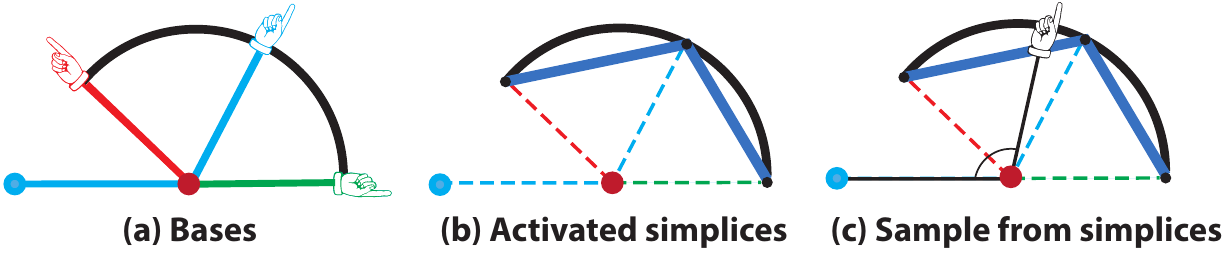}
\caption{Poses as the elbow is articulated. (a) The hand moves along the black curve as the elbow is articulated. The shoulder (blue dot) and elbow (red dot) are fixed in the poses. Our method learns two extreme bases (red and green) and an intermediate basis (cyan). (b) shows two activated simplices (blue line segments). The data on the black curve will be represented by their projections on the simplices. (c) shows a pose sampled from an activated simplex. All poses on the activated simplices satisfy the appropriate range of motion constraints at the elbow.
}
\vspace{-1em}
\label{fig:teasing}
\end{figure}

The Activated Simplices representation has several advantages: (i) Each simplex provides a tight approximation of a local region of the data manifold.  (ii) Curved and closed manifolds can be represented using an arrangement of simplices. (iii) The simplices can have different dimensions allowing the representation of an inhomogeneous data source; see Figure \ref{fig:ribbonCircle}. While the total number of bases  is a parameter of the model, the number and dimensions of the individual simplices are learned. (iv)  The representation is bounded.  This reduces the risk of generating data which violate constraints in certain applications. See Figure \ref{fig:teasing}. (v) Generative models can be constructed on the simplices using Dirichlet distributions which can be used to synthesize realistic data. (vi) The model is interpretable: points on the simplices are convex combinations of the vertices, and extremes among the data can be found among the vertices. See Figure \ref{fig:face_illumination} and \ref{fig:pose_basis}.

The requirement that data lie on the unit sphere may seem restrictive, however many data sources are normalized in that way, and we present some alternatives to the standard normalization in the Appendix.


The paper is organized as follows. Section \ref{sec:related_work} reviews related work. Sections \ref{sec:activated_simplices} and \ref{sec:pruning_activated_simplices} describe the activated simplices learning and pruning methods, and Section \ref{sec:optimization} discusses optimization strategies. The remainder of the paper is devoted to experiments on digits and poses, followed by a brief conclusion. In the Appendix, we discuss some projections onto the sphere which are useful as alternatives to the usual normalization, we present several low dimensional examples, we discuss the geometry of the standard sparse coding model, and we discuss our method in the context of methods for triangulating manifolds.

\section{Related Work}
\label{sec:related_work}
Our method represents data as a mixture of activated simplices. So it is natural to make comparisons with methods which represent data using a mixture of simple structures,  such as $k$-means, $k$-flats \cite{Canas:2012vf} and its generalizations and improvements including Sparse Subspace Clustering \cite{Elhamifar:2013uz}  and Atlas learning  \cite{Pitelis:2013ik}. Since our method builds on a convex hull, it is also natural to compare with Archetypal Analysis \cite{cutler1994archetypal, chen2014fast} which models the data globally as a convex hull and learns convex extremes.

$k$-means is a clustering algorithm where the data in each Voronoi region is represented by its mean. $k$-flats extends $k$-means  by using a vocabulary of hyperplanes, usually of the same dimension. The number of hyperplanes and their dimensions are parameters of the model. $k$-flats is limited in its ability to model curved manifolds, as it must use many hyperplanes for accurate reconstruction.

Sparse Subspace Clustering\cite{Elhamifar:2013uz} analyses the {\em self expression} (i.e., sparse linear interrelations) of the data and groups them according to hyperplanes. Atlas \cite{Pitelis:2013ik} learns hyperplane charts by fitting hyperplanes, with the restriction that an entire local neighbourhood of each point can be represented by the same hyperplane. This improves the assignment of points to hyperplanes at points of ambiguity. These methods give good performance when the data source resembles a small collection of hyperplanes. But they are limited when modelling data from a curved manifold, and the structures learned are not naturally bounded.

\begin{figure}
\centering
\begin{tabular}{cc}
\includegraphics[scale =0.23]{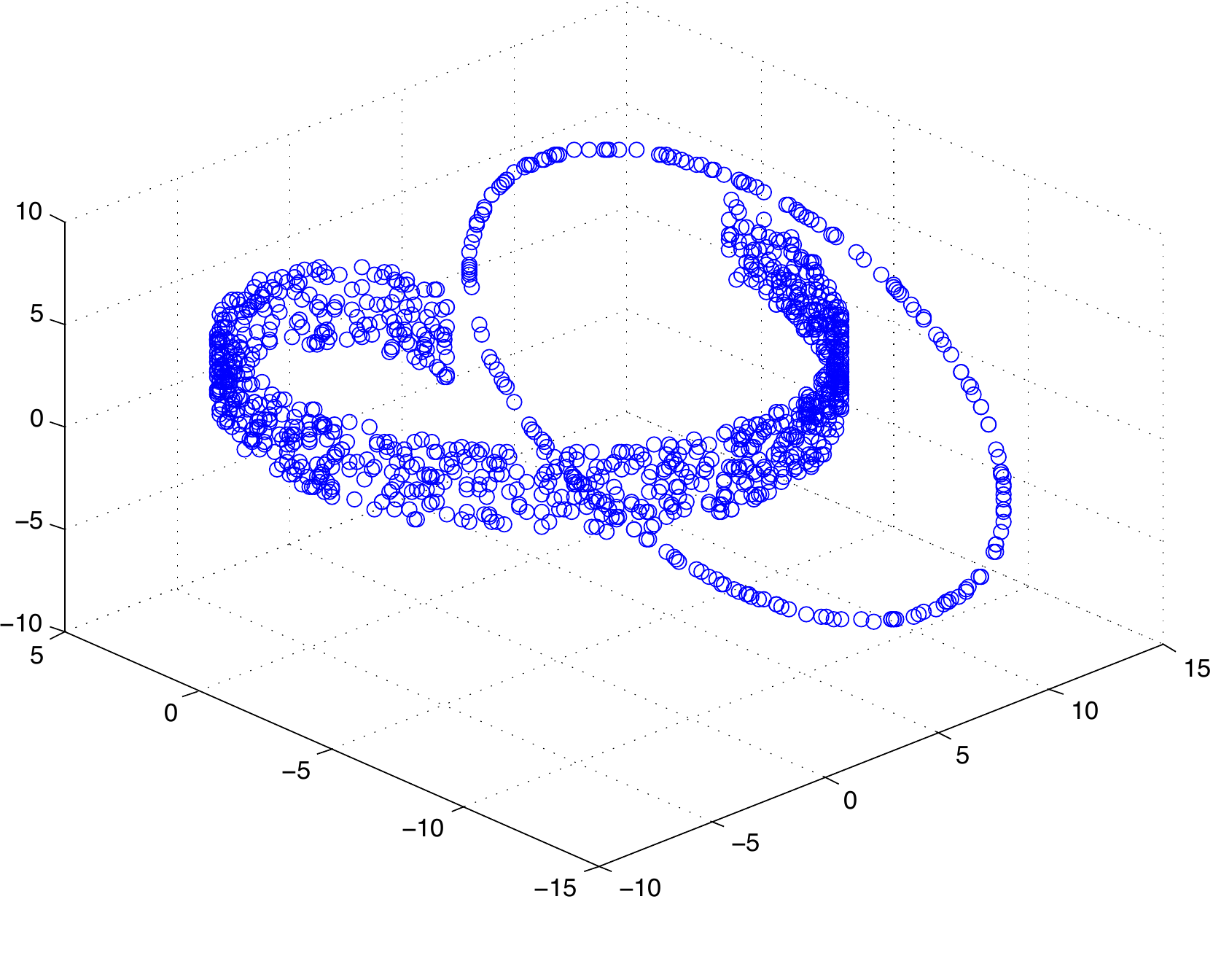}
\includegraphics[scale =0.21]{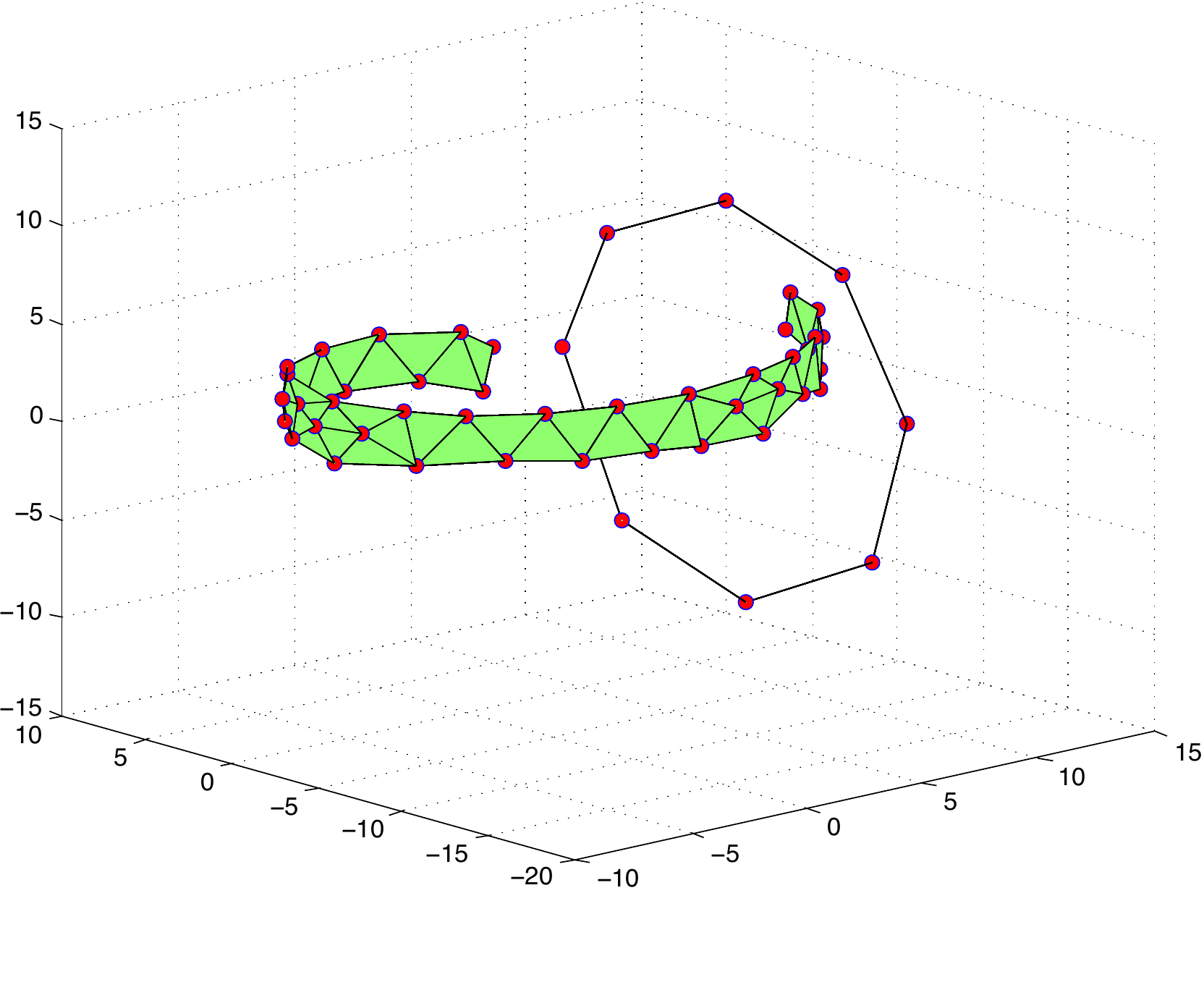}
\end{tabular}
\caption{The left panel shows the data points from the manifold of a ribbon passing through a circle. The right panel shows the detected basis vectors (red dots) and the activated simplices (line segments in the circle and the small triangles in the ribbon). The simplex dimensions can be different.}
\label{fig:ribbonCircle}
\vspace{-1em}
\end{figure}

Our simplicial model is more flexible in its vocabulary, since it assembles an arrangement of simplices to fit the data. It can manage curved manifolds and inhomogeneous sources (see Figure \ref{fig:ribbonCircle} and the Appendix for some pictures). It naturally accommodates constraints such as anatomical constraints, and allows for generative models by fitting a Dirichlet density to each simplex. It allows classification by nearest neighbour classification on the simplices. It will be preferable to hyperplane methods when the data is not close to a union of hyperplanes.

Archetypal Analysis \cite{cutler1994archetypal, chen2014fast} learns a global convex model for the data. It naturally discovers convex extremes. However when the data manifold is not a convex set, it ignores the details in the interior. The Activated Simplices method constructs local convex structures to describe the data. While it does construct a convex hull to approximate the data on the sphere, it is the boundary simplices of this hull, not the convex hull itself that are important.


\section{A Mixture of Simplices Model}
\label{sec:activated_simplices}

In our method we represent data using a mixture of simplices. Simplices are points, line segments, triangles, tetrahedra, and other higher dimensional analogs of the triangle; see Figure \ref{fig:simplex}. A point on a $k$-dimensional simplex with vertices $x_1, \ldots, x_{k+1}$ is a convex combination $\alpha_1 x_1 + \alpha_2 x_2 + \cdots + \alpha_{k+1} x_{k+1}$, where $\alpha_i \ge 0$ and $\alpha_1 +\alpha_2 + \cdots + \alpha_{k+1} =1$.
A simplicial model built using $p$ bases $\{x_1, \ldots, x_p\}$  is a family of simplices ${\cal F} =\{ \Delta_1 ,\ldots \Delta_{|{\cal F}|}\}$ each of whose vertices is a subset of the bases $X =\{x_1, \ldots, x_p\}$. Learning a simplicial model from data needs to solve the complex problem of learning the bases  $\{x_1, \ldots, x_p\}$, learning the simplices ${\cal F} =\{ \Delta_1 ,\ldots \Delta_{|{\cal F}|}\}$, learning the assignment of each data point to a simplex, that is, picking a simplex for each data point, and learning the projection of each data point on its chosen simplex. For training data $Y=\{y^{(1)}, \ldots y^{(N)\}}$ this is the following minimization:

\begin{equation}
 \min_{X, {\cal F}, \alpha, V} \  \frac{1}{N}\sum _j^N \sum_{t =1}^ {|{\cal F}|} V_{t}^{(j)} \| y^{(j)} - \Delta_t \alpha_t ^{(j)}\|^2
\end{equation}
where $V$ is a binary indicator variable assigning each point to exactly one simplex in the family, each $\alpha_t ^{(j)}$ is a convex coefficient vector that attempts to express data point $y^{(j)}$ in terms of the $t^{{\text th}}$ simplex $\Delta_t$.

We now assume that the data are normalized to have $\ell_2$ norm 1 and hence lie on a unit sphere, and that the bases have $l_2$ norm at most 1, so that they are comparable to the data. This simplifies the assignment problem dramatically, since the data now lie on the surface of a sphere, and the simplices $\Delta_i$ are convex combinations of bases within the sphere.  It simplifies the problem of identifying optimal simplices, since the convex combinations of the bases that are closest to the data are on boundary facets of the convex hull of the bases. The optimal construction of $\cal F$ from the bases $X$ is the set of boundary facets activated in this way by the data.  Since facets of a convex hull in general positions are simplices, we call these facets activated simplices. The optimal coefficients $\alpha_t^{(j)}$ are those from the projection of the data point $y^{(j)}$ onto the boundary of the convex hull.

We've observed that the best projection of $y^{(j)}$ onto the optimal simplices is the projection of $y^{(j)}$ onto the convex hull of the bases. This means that  in the optimization
\begin{equation}
\min_{\beta^{(j)}}\|y^{(j)} -X\beta^{(j)}\|^2
\end{equation}
subject to the constraints $\beta^{(j)} \ge 0$, $\|\beta^{(j)}\|_1 =1$, which expresses $y^{(j)}$ as a convex combination of {\bf all} the bases, the non-zero entries in the coefficient vector $\beta^{(j)}$ identify the boundary simplex on which $y^{(j)}$ is projected. Thus we can learn a mixture of simplices model by the following minimization

\begin{equation}
\label{eq:newformulation}
\min_{X, \beta} \frac{1}{N} \sum _{j=1}^N ||y^{(j)} -X\beta^{(j)}||^2
\end{equation}
subject to $\beta^{(j)} \ge 0$, $\|\beta^{(j)}\|_1 =1$, for $j = 1, \ldots, N$, and $\|x_i\|_2 \le 1$ for $ i = 1, \ldots, p$.

We then derive the activated simplices by studying the activations $\beta^{(j)}$.
\begin{figure}
\centering
\includegraphics[height=1in]{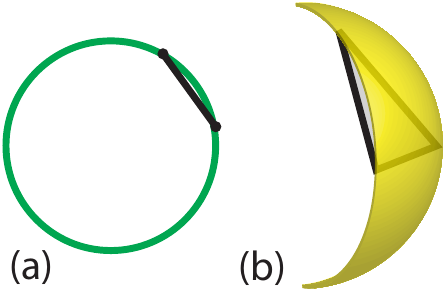}
\caption{Low dimensional facets are closer to the sphere than high dimensional ones. (a) The interior of the line segment (1-simplex) is further from the circle than its endpoints (0-simplics). (b) The edges of the triangle are closer than its interior.}
\label{fig:low_dimensional_motivation}
\vspace{-1em}
\end{figure}

\subsection{The Activated Simplices Model}

To summarize the observations above, we develop a mixture of simplices model for data $Y = \{y^{(1)}, \ldots, y^{(N)}\}$ which is assumed to be unit normalized. The simplices are learned from co-activated bases in the optimization (\ref{eq:newformulation}). The optimization  (\ref{eq:newformulation}) can be understood as learning a convex hull $C_X$ of the bases $X$, such that its boundary facets are close to the training data. We call the boundary facets that are closest to the data {\bf activated simplices} --- these can be determined from activations of the coefficients $\beta^{(j)}$.

The collection of simplices that are learned from (\ref{eq:newformulation}) can be pruned to obtain a more efficient representation. This is discussed in Section \ref{sec:pruning_activated_simplices}.

 \subsection{Activated Simplices are Low Dimensional}

 We find in experiments that the activated simplices  have low dimension. However the boundary facets for a generic convex body in $d$-dimensional space are $d$ dimensional simplices. Our method learns facets (that is, sub-simplices of boundary simplices) that have dimension much less than $d$. We now give an explanation for this phenomenon.

Suppose that the data $Y$ are from a low dimensional manifold embedded in $R^d$. Recall that in the minimization (\ref{eq:newformulation}), a basis $X$ is learned so that boundary facets of the convex hull $C_X$ are close to the data. The learning process constructs $X$ so that this representation is as efficient as possible.

The interaction of the curved geometry  of the sphere with the linear geometry of the boundary facets of $C_X$, means that it is most efficient to position low dimensional facets close to the data.
Figure \ref{fig:low_dimensional_motivation} shows a segment (a 1-simplex) with its vertices in the circle, and it shows  a triangle (a 2-simplex) with its vertices in the 2-sphere. Notice that the interior of the segment is further from the sphere than its endpoints, similarly the interior of the triangle is further from the sphere than its boundary segments. In general, it is more efficient to approximate points on the sphere locally with low dimensional facets, than with high dimensional facets. Hence the optimization (\ref{eq:newformulation}) learns a structure that approximates the data with low dimensional facets of $C_X$.

Of course if the data is distributed uniformly over the sphere, it will be impossible to position low dimensional facets close to the data points, without a vast number of bases. But we see in our experiments, with what are understood to be low dimensional data sources, that the activated simplices are low dimensional. This makes sense as it provides a most efficient representation of the data.

\begin{figure}
\includegraphics[scale =0.65]{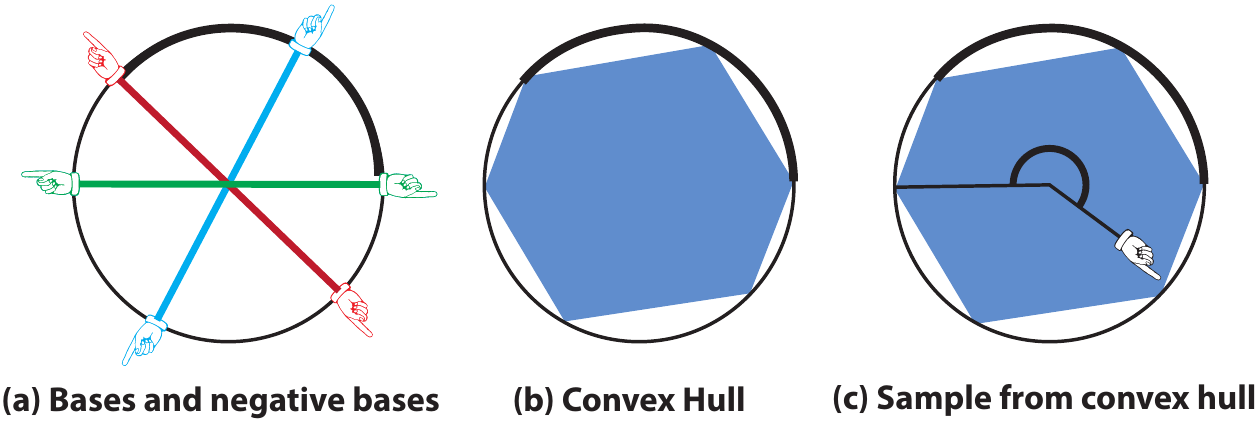}
\caption{Learning bases using standard sparse coding. (a) Sparse coding represents data using combinations of bases and negative bases. (b) In the usual prior learned from sparse coding, any combination of bases with small $\ell_1$ norm is allowed. Here we show combinations with $\ell_1$ norm at most one, which corresponds to the convex hull of the bases and negative bases. (c) A sample from this convex hull clearly violates the range or motion constraints at the elbow. The activated simplices in Figure \ref{fig:teasing} are a better prior.
}
\vspace{-1em}
\label{fig:teasing_2}
\end{figure}

\begin{figure*}
\includegraphics[height=1.3in]{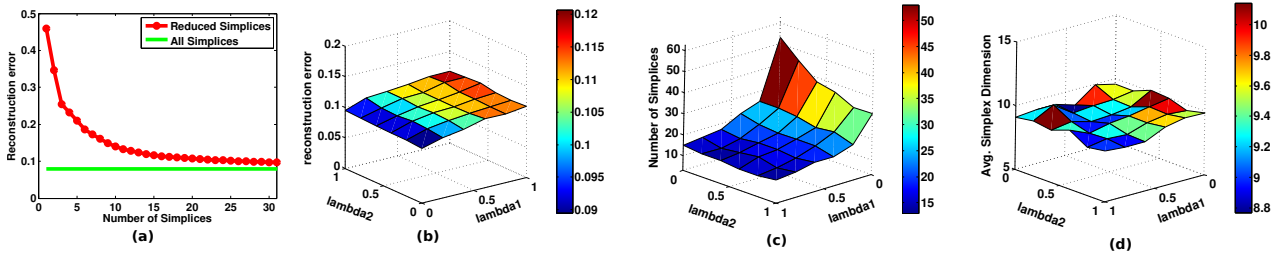}
\caption{\textbf{(a)} The effect of simplices pruning. The x-axis is the number of the selected simplices. The y-axis is the corresponding reconstruction error. The number of pre-pruned simplices is about $3,000$. The method obtains comparable reconstruction errors by selecting about $30$ simplices from the $3,000$. \textbf{(b,c,d)} show the influence of the two parameters $\lambda_1$ and $\lambda_2$ on the reconstruction error, the number of simplices and the average dimensions of the simplices.}
\label{fig:pose_prune}
\end{figure*}

\subsection{Relation to Sparse Coding, and a Penalized Version of Activated Simplices}

The coefficients vectors $\beta^{(j)}$ learned in (\ref{eq:newformulation}) are sparse, that is, most entries are zero, and the non-zero entries identify a low dimensional boundary facet of the convex hull of the bases $X$. Some readers will be reminded of sparse coding. Indeed it is possible to understand sparse coding using similar ideas --- we discuss this in the Appendix.

In the usual formulation of sparse coding \cite{tibshirani1996regression, Osborne:2000ha, Bradley:2004tz, Huggins} data $Y$ are represented by bases $X$ through the following $\ell_1$ penalized optimization:
\begin{equation}
\label{eq:standardsparsecoding}
\min_{X, \beta} \frac{1}{N} \sum _{j=1}^N \|y^{(j)} -X\beta^{(j)}\|^2  + \lambda \|\beta^{(j)}\|_1
\end{equation}
The bases $x_i$ are constrained to have $\ell_2$ norm at most 1.
Notice there is no constraint here that the coefficients $\beta^{(j)}$ are positive. The method learns sparse representations of the data as combinations of the bases and negative bases.

The bases learned in sparse coding can be used as a regularizer for pose data, in a similar spirit to our experiments.  See \cite{Wang:2014kv}. However the usual prior associated with a sparse coding representation, is that any combination $X\beta$ can occur, provided that $\|\beta\|_1$ is small. This allows co-activations of the bases that don't occur in the training data. Figure \ref{fig:teasing_2} shows a regularizing structure that might be learned in this way. Compare with Figure \ref{fig:teasing}, to see that the Activated Simplices method learns a more realistic representation.

It is interesting to experiment with a penalized version of the Activated Simplex method. A formulation with positive penalty parameter $r \le 1$ is
\begin{equation}
\label{eq:spenalizedactivatedsimplices}
\min_{X, \beta} \frac{1}{N} \sum _{j=1}^N \|y^{(j)} -X\beta^{(j)}\|^2
\end{equation}
subject to $\beta^{(j)} \ge 0$, $\|\beta^{(j)}\|_1 =r$, for $j = 1, \ldots, N$,
 and $\|x_i\|_2 \le 1$ for $ i = 1, \ldots, p$.

 This will force representations on lower dimensional facets when $r<1$. We've experimented with this penalized version and found that the unpenalized version works well with the data we considered. But we can imagine that the additional sparsity provided in (\ref{eq:spenalizedactivatedsimplices}) might be useful in managing some high dimensional data.

\section{Pruning The Activated Simplices}
\label{sec:pruning_activated_simplices}
Each training point $y^{(j)}$ proposes (activates) a simplex $\Delta$. Gathering the activated simplices, we obtain a set of simplex proposals ${\cal F} =\{ \Delta_1 ,\ldots \Delta_{{ T}}\}$. The number of proposed simplices will typically be much less than the number of training data, since several data points will activate the same simplex.

To prevent over-fitting, we propose a principled way to prune the simplices $\cal F$ to a proper subset $\cal F^*$ without degrading the reconstruction results much.  For example, $\cal F$ might contain a high dimensional simplex $\Delta$ that received few activations, but its low dimensional sub-simplices might have received many activations. We may obtain a better model if we remove this large simplex from the model.

We prune so as to obtain a small number of simplices  and so that the remaining simplices have low dimensions. We use the following to balance the goals of few and low dimensional with the goal of  good reconstruction:
\begin{equation}
\begin{split}
\cal{F}^* &= \argmin_{\hat{\cal{F}} \subset \cal{F}}{L(Y, \hat{\cal{F}})+\lambda_1 \#(\hat{\cal{F}})+\lambda_2 \sum_{i=1}^{|\hat{\cal{F}}|}{\dim(\hat{\cal{F}}_i)}},
\end{split}
\label{eq:prune}
\end{equation}
where $L(Y,\hat{\cal{F}})$ represents the reconstruction error using the set  $\hat{\cal{F}}$ of simplices to approximate the data, and $\#(\hat{\cal{F}})$ is the number of simplices in the set $\hat{\cal{F}}$, and $\dim(\hat{\cal{F}}_i)$  is the dimension of the $i^{\text{th}}$ simplex in the set $\hat{\cal{F}}$. The two penalty terms $\lambda_1 \#(\hat{\cal{F}})$ and  $\lambda_2 \sum_{i=1}^{|\hat{\cal{F}}|}{\dim(\hat{\cal{F}}_i)}$ encourage fewer simplices and encourage lower dimensions. We set $\lambda_1$ and $\lambda_2$ to be $0.001/T$ and $0.01/T$, respectively, for our experiments. In general, these can be set by cross-validation. Figure \ref{eq:prune} shows the influence of pruning.

\begin{figure}
\centering
\includegraphics[height=1.8in]{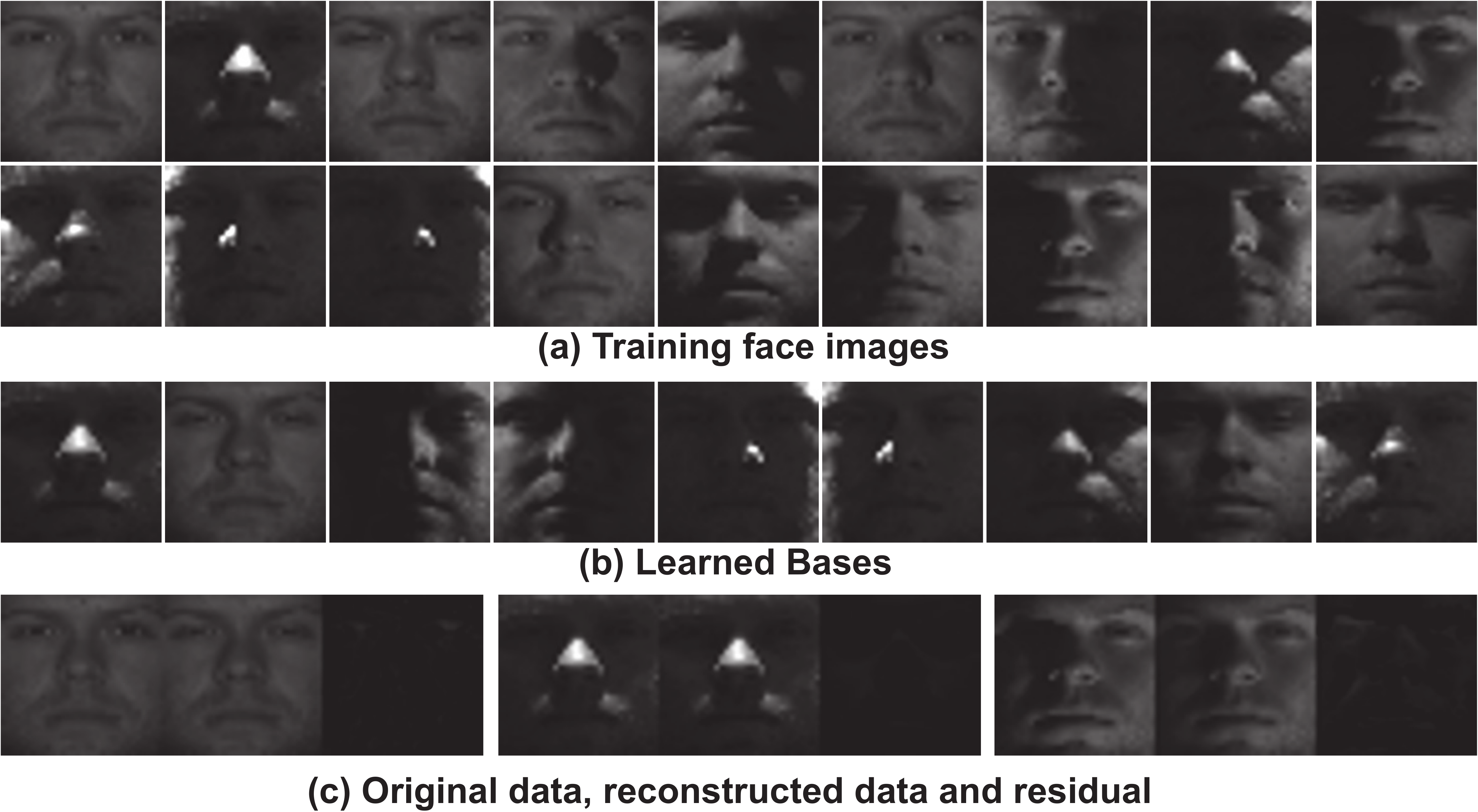}
\caption{The top row shows sample faces of different lighting directions in terms of Azimuth and Elevation in the Yale Face B dataset. The mid row shows the learned bases. We can see that some of them approximately correspond to the faces of extreme lighting conditions. The bottom row shows the reconstructed faces and residuals. }
\label{fig:face_illumination}
\end{figure}

\section{Optimization}
\label{sec:optimization}

The formulation (\ref{eq:newformulation}) is non-convex, but it is convex with respect to each of the variables, $X$ and $\beta$, when the other is fixed. Optimizing the two convex sub-problems alternately leads to a local minimum, but using classic convex optimization methods may be slow when the number of training data is large. Inspired by \cite{Mairal:2009um}, we solve the problem in an online way, based on stochastic gradient descent. This scales up gracefully to large dataset. When we fix $\beta$ and optimize the bases $X$, we obtain the same dictionary update problem as in \cite{Mairal:2009um}. So we refer the readers to \cite{Mairal:2009um} for more information. When we fix $X$ and update $\beta$, we obtain a least-squares problem with a simplex constraint.  We use an active-set algorithm \cite{wright1999numerical} to benefit from the sparsity of $\beta$.

The problem (\ref{eq:prune}) is difficult to optimize directly because of the large search space. Hence we use a greedy approach which successively  adds simplices to $\hat{\cal F}$ which  most reduce the objective (\ref{eq:prune}). We terminate the process when the objective stops decreasing.
On a regular Desktop equipped with $3.4$GHz CPU, the overall simplices learning time is about $0.19$ seconds for a digit dataset containing $800$ training data of dimension $256$. Projecting data on the simplices is fast.

\begin{figure}
\centering
\begin{tabular}{cc}
\includegraphics[scale =0.13]{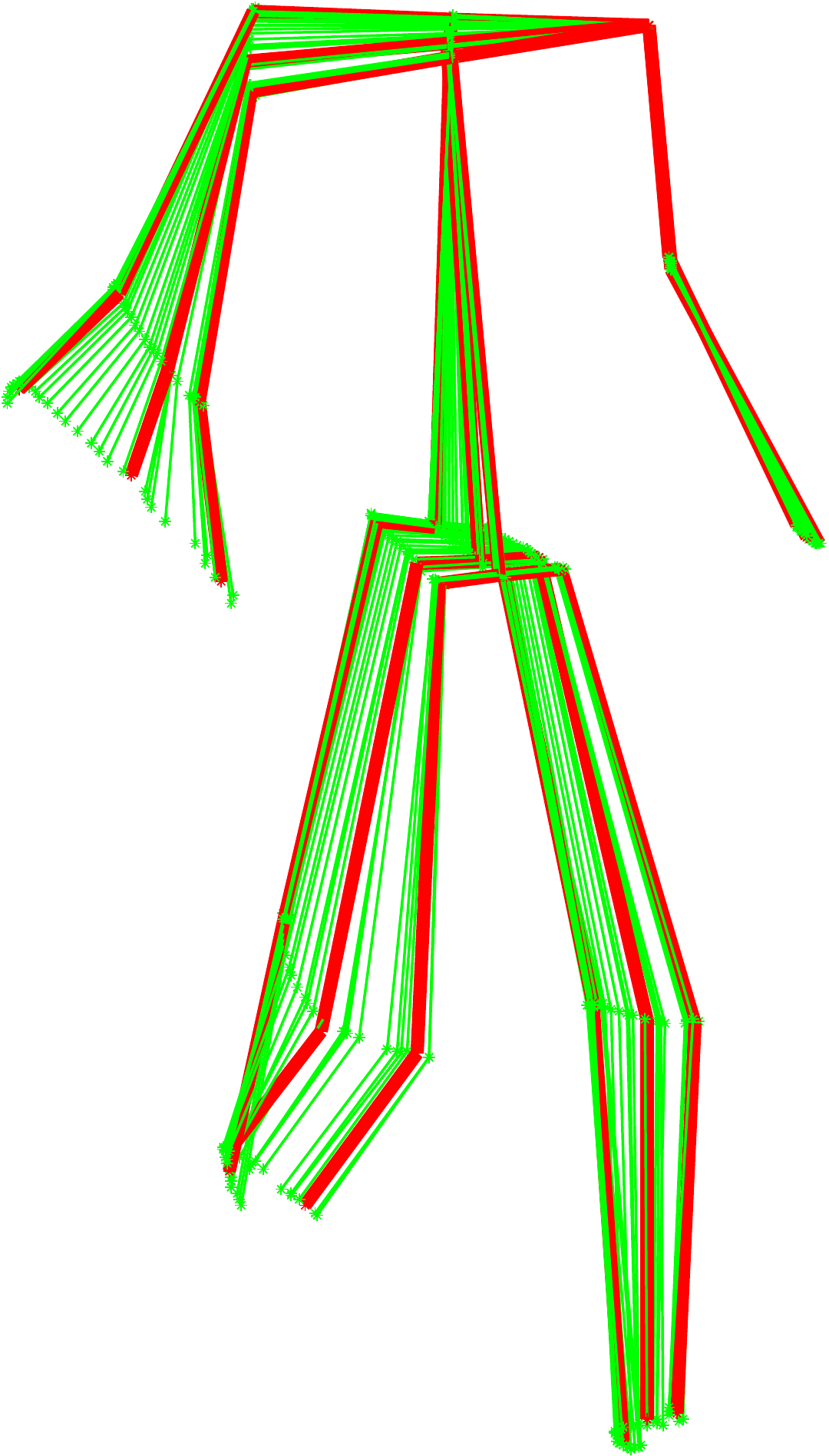}
\includegraphics[scale =0.13]{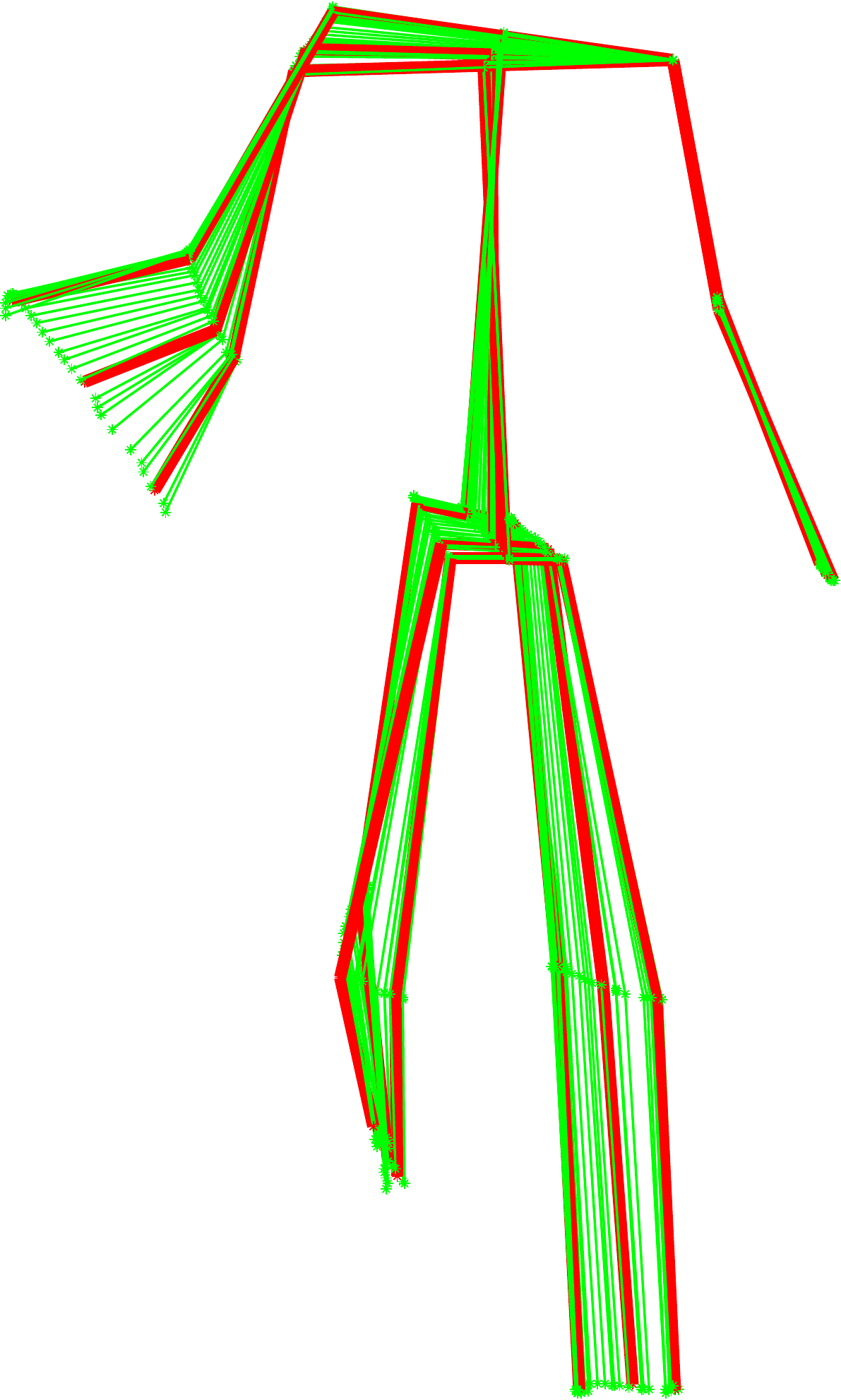}
\end{tabular}
\caption{The figure shows $30$ human poses from a walking sequence (in green) and the learned three bases (in red) overlaid together, viewed from two different directions. We can see that our method actually learns three extreme bases.}
\label{fig:pose_basis}
\vspace{-1em}
\end{figure}

\section{Data Reconstruction}
\label{sec:reconstruction_experiment}
In this section we evaluate the reconstruction errors by computing the Euclidean distance between data and their projections on the nearest simplices. We evaluate the influence of the model parameters $p$, i.e., the number of bases. We also show some results when the convex hull radius $r$ is smaller than one.

We conduct experiments on a large human pose dataset H3.6M \cite{ionescu2013human3} which contains $3.6$ million poses. We select $11,000$ poses of $11$ actions including ``walking'', ``taking photo'', ``smoking'', ``sitting down'', ``sitting'', ``purchases'', ``posing'', ``phoning'', ``greating'', ``eating'' and ``discussion''.

\begin{figure}
\centering
\begin{tabular}{cc}
\includegraphics[scale =0.26]{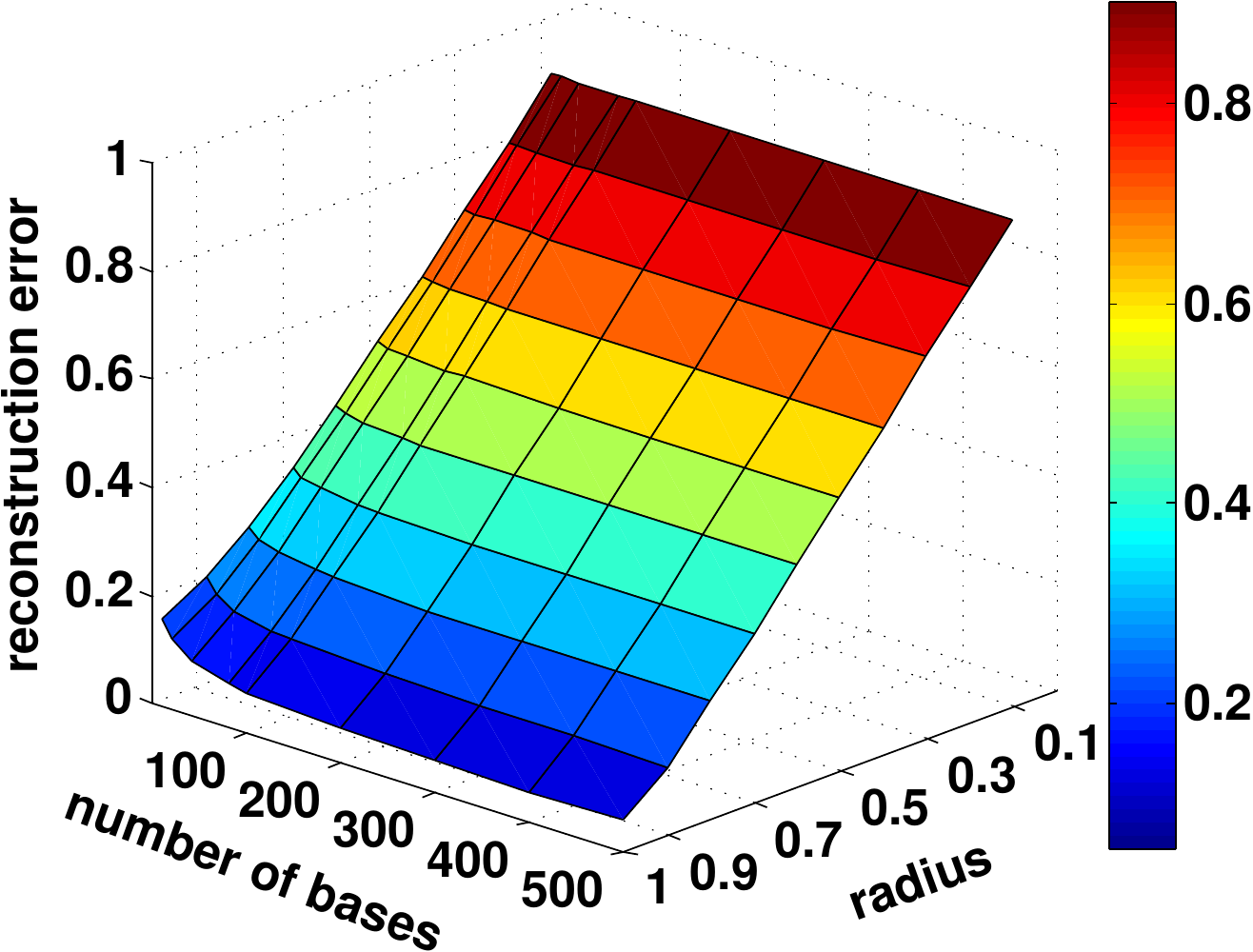}
\includegraphics[scale =0.26]{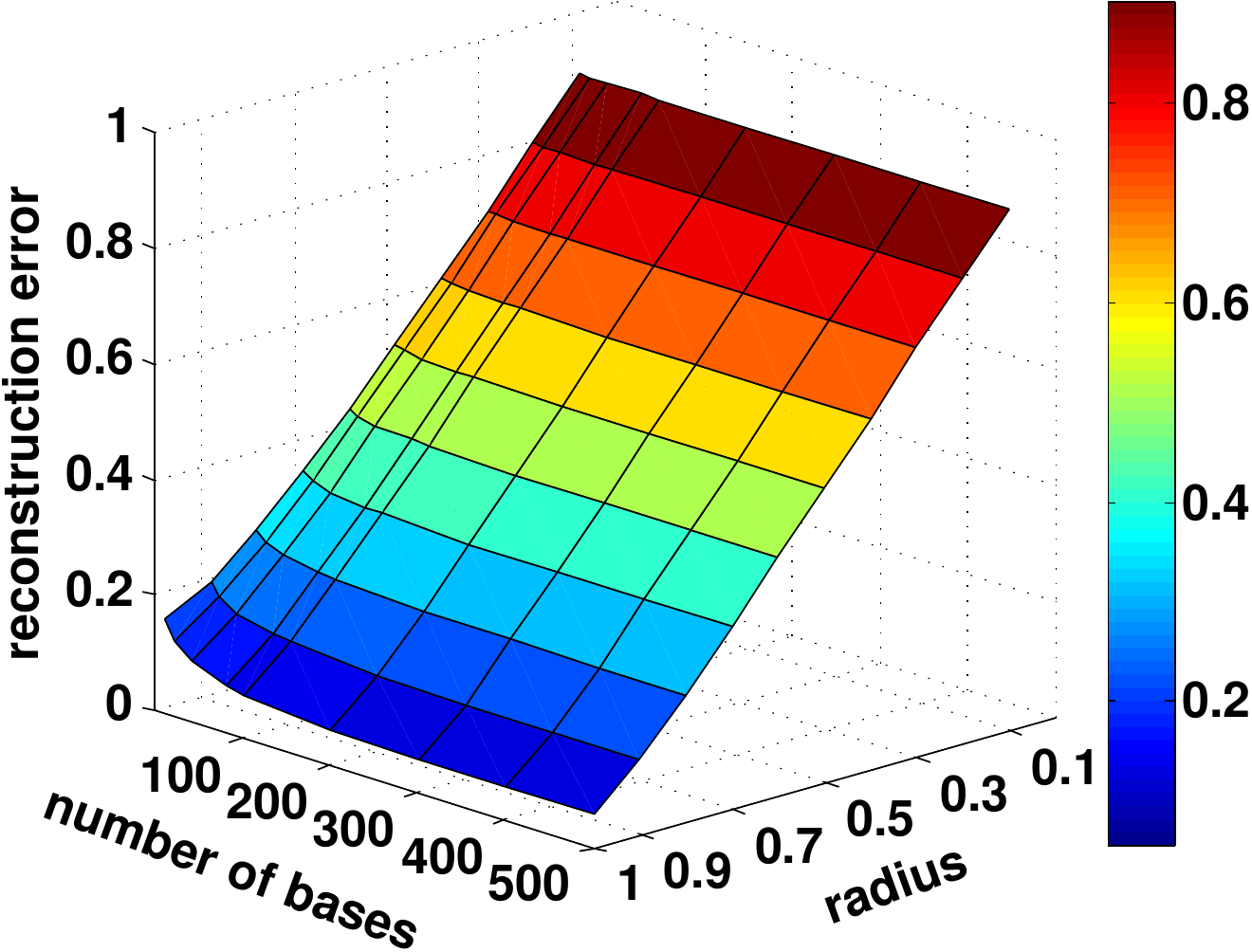}
\end{tabular}
\caption{Reconstruction errors in terms of the number of bases and the scale $r$ of the convex hull. The left figure shows the reconstruction error when training and testing on the same actions in the dataset. We can see that the least reconstruction error (achieved when $r=1$ and $p=500$) is almost zero which justifies the activated simplices approximate the data well. The right figure shows the reconstruction errors when training and testing on different actions.}
\label{fig:pose_reconstruction_quan}
\end{figure}

We first evaluate on the training data--- we train and test on the same $11,000$ poses. We learn a single set of simplices for the eleven actions together. Figure \ref{fig:pose_reconstruction_quan} (left) shows the average reconstruction errors for each choice of $p$ and $r$. The smallest average reconstruction error is about $0.0423$ (achieved when $r=1$ and $p=500$) which indicates that the activated simplices can represent the training data well. The reconstruction errors decrease fast as radius $r$ increases. The reconstruction errors also decrease as the number of bases $p$ increases but in a more stable manner.

\begin{figure}
\centering
\begin{tabular}{ccc}
\includegraphics[scale =0.3]{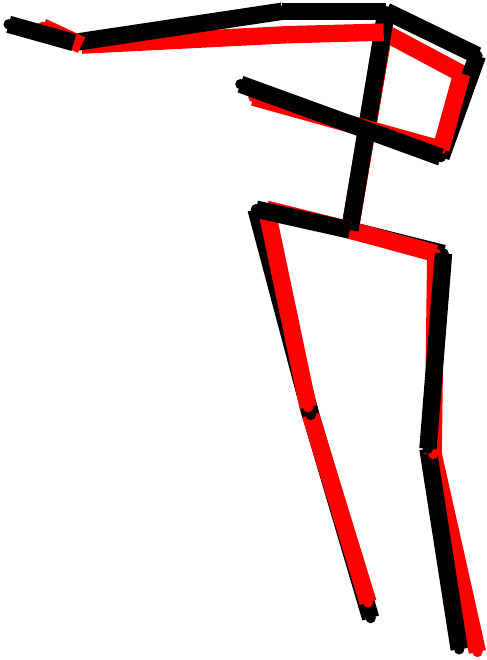}
\includegraphics[scale =0.28]{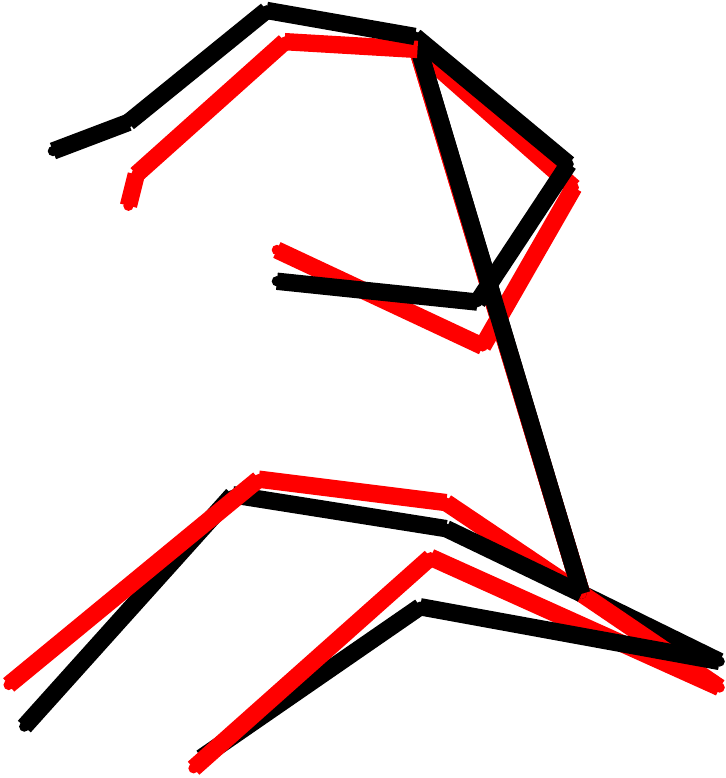}
\includegraphics[scale =0.28]{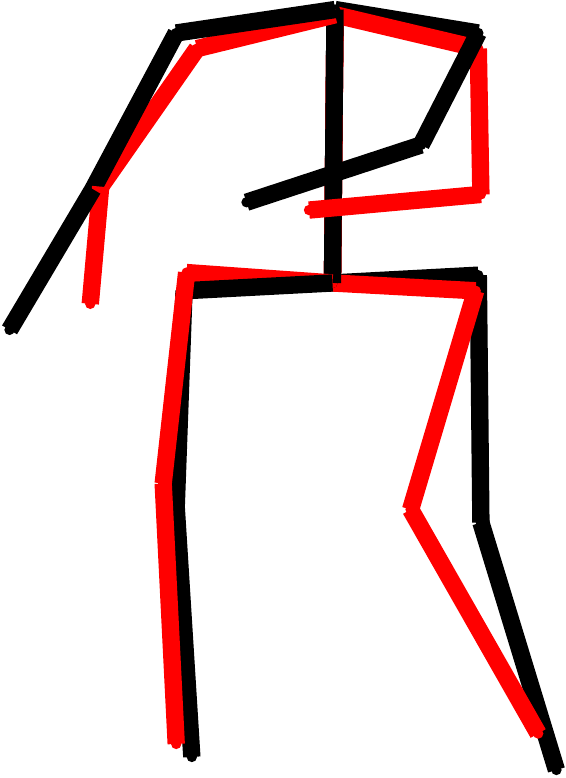}
\end{tabular}
\caption{Sample reconstruction results. The poses in black are the groundtruth poses and the red ones are their projections on the activated simplices. The reconstruction errors are $0.06$, $0.1$ and $0.2$, respectively, from left to right.}
\label{fig:pose_reconstruction_qual}
\end{figure}

In the second scenario, we split the $11,000$ poses into training and testing subsets each containing $5,500$ poses of the $11$ actions. We learn the activated simplices from the training data and test on the testing data. Figure \ref{fig:pose_reconstruction_quan} (right) shows the results. We can see that the reconstruction error is similar to that of the first experiment. The smallest reconstruction error is about $0.05$ (achieved when $r=1$ and $p=500$). The results justify that the activated simplices model can reconstruct the data well. Figure \ref{fig:pose_reconstruction_qual} shows three sample reconstruction results whose reconstruction errors are $0.06$, $0.1$ and $0.2$, respectively.

\section{Nearest Neighbour Classification}
\label{sec:classification}
We conduct digit classification on the Semeion handwritten digit dataset \cite{semeion_dataset} and action classification on a more challenging MSR-Action3D dataset \cite{li2010action}. We use simple nearest neighbour classifiers on the simplicial model and achieve the state-of-the-art performance on both tasks.

\subsection{Handwritten Digit Classification}
The Semeion dataset \cite{semeion_dataset} contains $1593$ grey scale images of handwritten digits ($0,\ldots,9$) generated from about 80 people who each wrote all ten digits twice. The images are of size $16 \times 16$ and each pixel in the image is converted into a boolean value using a fixed threshold. We randomly select $796$ images for training and $797$ images for testing following \cite{Pitelis:2013wg}. The process is repeated for $100$ times and the average result is reported. We compare with the standard manifold learning methods such as LTSA \cite{Zhang:2004ht}, Atlas \cite{Pitelis:2013ik}, Archetypes \cite{chen2014fast} and the usual sparse coding.

\begin{table}
\caption{Digit classification accuracy (\%) of sparse coding, LTSA \cite{Zhang:2004ht}, Atlas \cite{Pitelis:2013ik}, Archetypal Analysis \cite{chen2014fast} and our approach. The best performance of different dimension choices is reported for LTSA and Atlas. The dimension of our method is automatically determined by the algorithm.} \centering
\begin{tabular}{|c|c|c|}
\hline
Methods & Best Dimensionality & Accuracy(\%) \\
\hline
Sparse Coding & 10 & 89.40 \\
\hline
LTSA \cite{Zhang:2004ht} & 34 & 90.44 \\
\hline
Atlas \cite{Pitelis:2013ik} & 31 & 91.73 \\
\hline
Archetypes \cite{chen2014fast} & 10 & 91.17 \\
\hline
\textbf{Our} & \textbf{4-8} & \textbf{93.00} \\
\hline
\end{tabular}
\label{table:digit}
\end{table}

In the training stage, for our method, we learn a set of activated simplices for each of the ten classes using the same parameters. In the testing stage, for each data, we project it to the ten classes of simplices independently and obtain the class having the smallest reconstruction error. We set the number of bases $p$ to be ten for each class ($100$ in total). We finally obtain $100$ simplices whose dimensions are from four to eight. For sparse coding and Archetypes methods, we learn a set of bases for each of the ten classes and project test data to the ten sets of bases independently and obtain the class which has the smallest error.

\begin{figure}
\centering
\begin{tabular}{cc}
\includegraphics[scale =0.25]{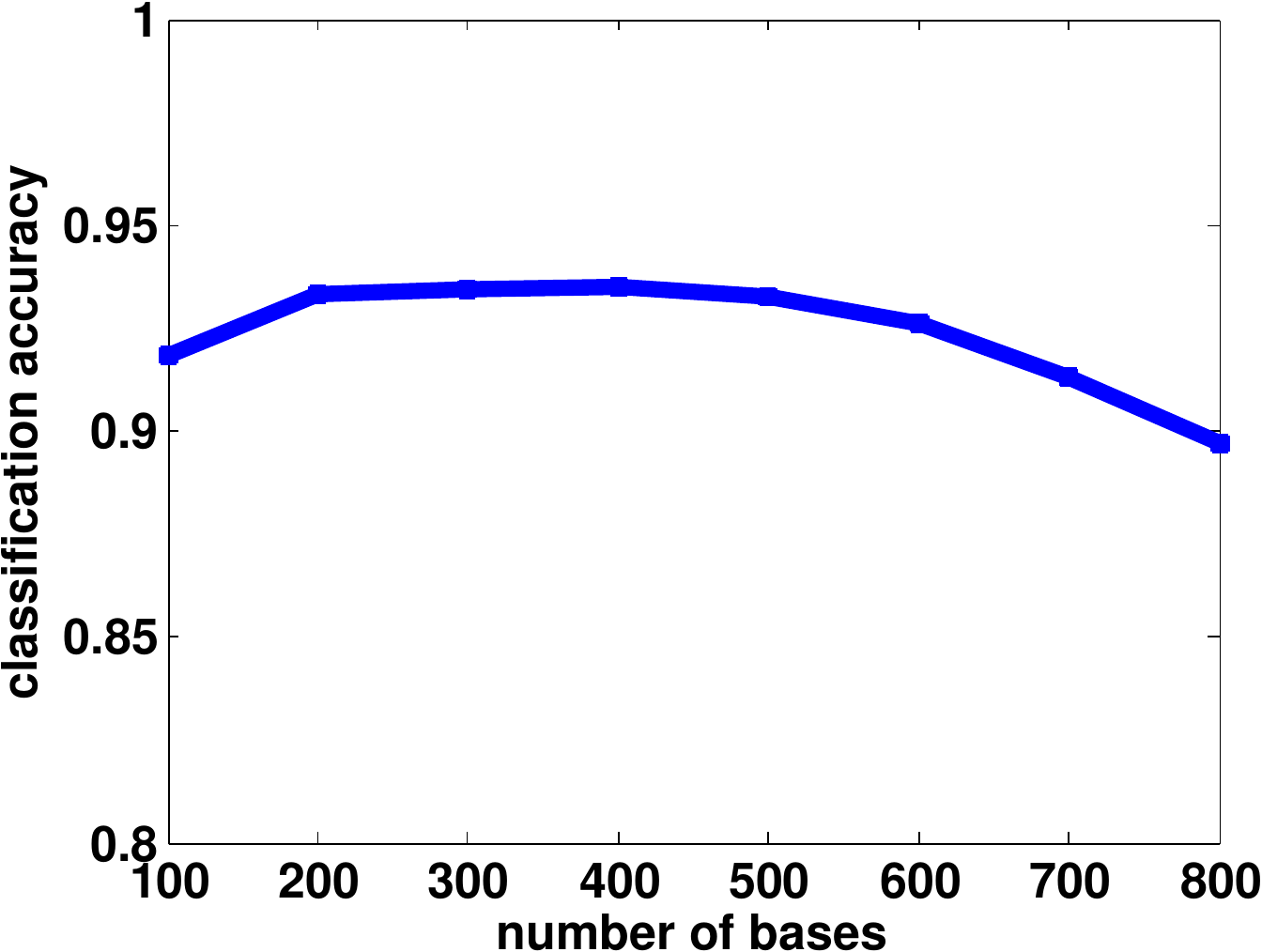}
\includegraphics[scale =0.25]{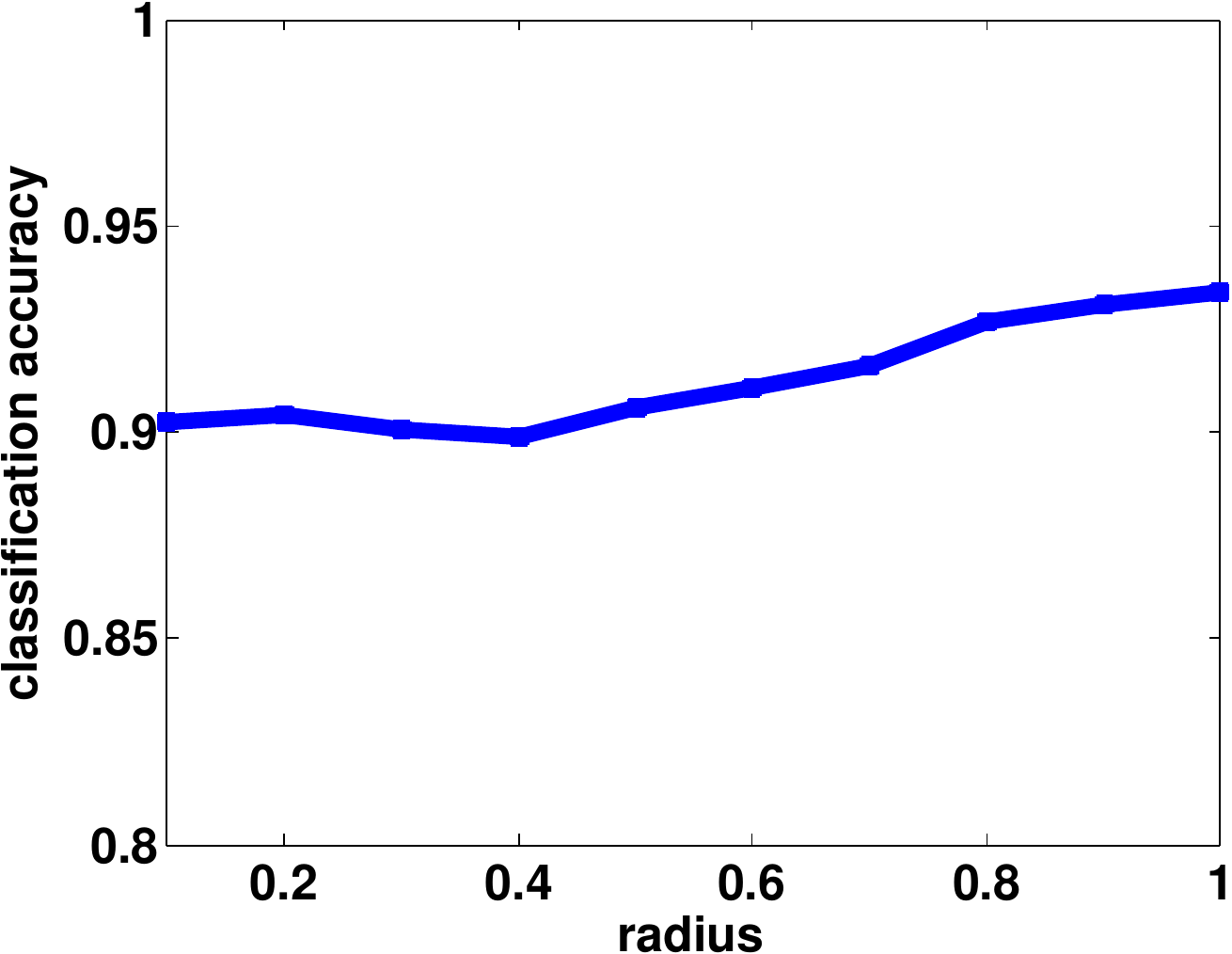}
\end{tabular}
\caption{Classification accuracy on the Semeion dataset using different numbers of bases and the convex hull radius. In the left figure, the classification accuracy begins to decrease when the number of bases is larger than $400$. This is because too large a number of base causes over-fitting.}
\label{fig:influence}
\vspace{-1em}
\end{figure}

Table \ref{table:digit} shows the recognition results. We outperform all of the methods. Our approach automatically determines the simplex dimensions which vary from simplex to simplex. For other methods, different choices of dimensions have been tried and the one that achieves the best performance is reported. The number of bases is $100$ for sparse coding and archetype for fair comparison. Figure \ref{fig:influence} shows the influence of the number of bases and the radius of the convex hull on the classification results.

\subsection{Action Recognition}

MSR-Action3D dataset \cite{li2010action} provides $557$ human pose sequences of ten subjects performing 20 actions which are recorded with a depth sensor. There are about $50$ frames in each sequence. This is a challenging dataset because many of the actions are highly similar to each other. We use the cross-subject evaluation scheme as in \cite{li2010action}.

To make the poses more discriminative between similar classes, for each pose in the sequence, we stack it with its consecutive ten poses as a high dimensional pose snippet. Weak temporal coherence can be achieved by this simple stacking operation. The activated simplices are learned based on snippets rather than a single pose.

In the training stage, for each of the 20 classes, we learn an independent set of activated simplices using the same parameters. In the testing stage, for each sequence, we project the 3D snippets onto the 20 sets of activated simplices and select the class which has the smallest reconstruction error. We set the number of bases for each class to be $25$ ($500$ in total) by cross-validation.

\begin{table}
\caption{Action recognition accuracy on MSR-Action3D.} \centering
\begin{tabular}{|c|c|}
\hline
Methods & Accuracy (\%) \\
\hline
Sparse Coding & 80.22 \\
\hline
Action Graph on Bag-of-3D joints \cite{li2010action} & 74.70\\
\hline
Actionlet Ensemble \cite{wang2012mining} & 88.20\\
\hline
Spatial-Temporal-Part model \cite{Wang:2013kv} & 90.22 \\
\hline
\textbf{Our Approach} & \textbf{91.30} \\
\hline
\end{tabular}
\label{table:msr_action3d}
\end{table}

Table \ref{table:msr_action3d} shows the results. Our method outperforms \cite{li2010action} and sparse coding, and achieves comparable performance as \cite{wang2012mining} and \cite{Wang:2013kv}. However, actionlet ensemble uses sophisticated feature learning methods and multiple kernel learning classifiers. Spatial-temporal-part model uses data mining techniques to remove the noisy joints and then use support vector machine for classification. In contrast, our method is the simplest in terms of both features and classifiers.

We have an interesting observation from the experiment. For each test sequence, if we select the top $K$ classes which have the smallest reconstruction errors and regard the classification is correct when any of the $K$ predictions is correct, then we can considerably improve the performance. For example, when $K=2$, the accuracy increases to $95.54\%$; when $K=5$, the accuracy increases to $97.03\%$; when $K=8$, the accuracy is $100\%$.

\section{3D Human Pose Estimation}
\label{sec:pose_estimation}
In this section, we apply the simplicial model to the task of 3D human pose estimation from 2D poses. This allows us to compare our method with the more standard application of sparse coding \cite{Wang:2014kv}.

The 3D and 2D poses are represented by $n$ joint locations $P \in R^{3 \times n}$ and $O \in R^{2 \times n}$, respectively.
We conduct experiments on the H3.6M Human motion capture dataset \cite{ionescu2013human3} which provides ground truth 3D human joint locations of $11$ actions. To evaluate the generalization ability of the method, we select the poses from the first five actions for training and the remaining six actions for testing.

We obtain the 2D joint locations by projecting the 3D pose into 2D using synthetic weak perspective camera parameters $M$, i.e., $O=M \cdot P$, where $ M=\left(
\begin{array}{c}m_1^T
\\ m_2^T\end{array}\right) \in \mathbb{R}^{2 \times 3}$. We assume the joint locations are mean-centered hence eliminate the translation component for simplicity. We now have 3D joint locations  $P$ which we use as ground truth, and we have the corresponding 2D joint locations $O$ --- our goal is to reconstruct $P$ from $O$ with the help of the activated simplices.

\begin{figure}
\centering
\includegraphics[scale =0.28]{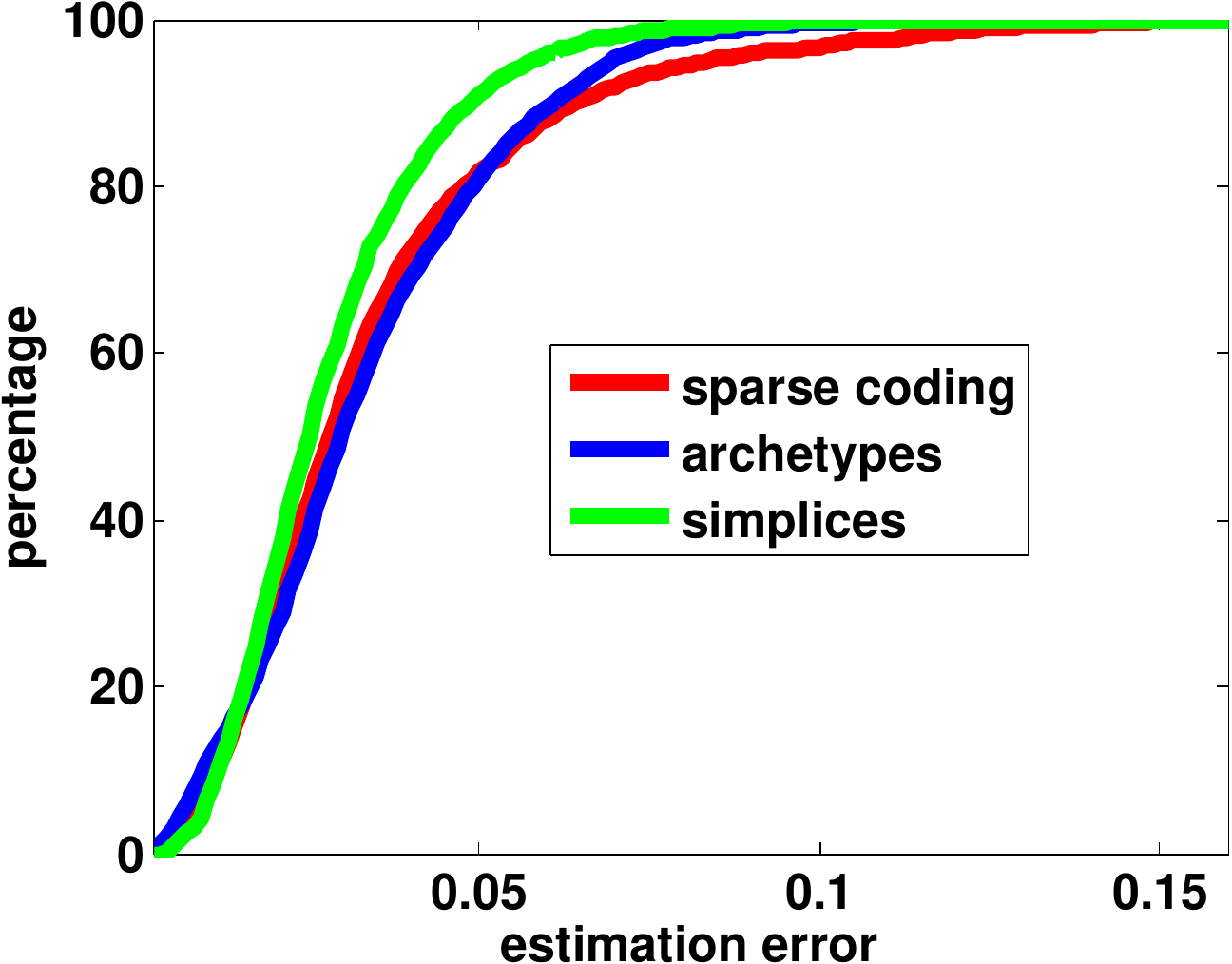}
\includegraphics[scale =0.28]{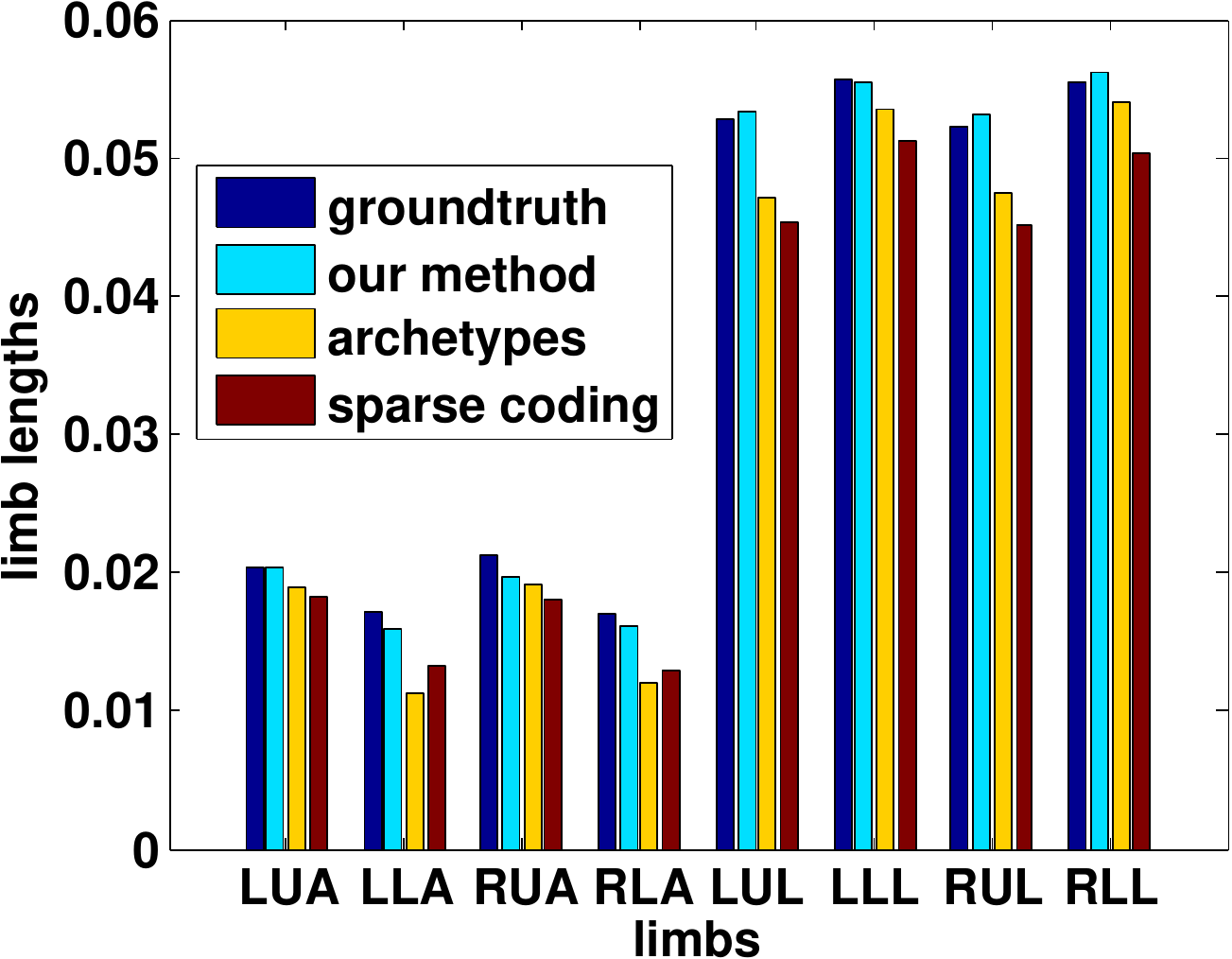}
\caption{Left figure compares the estimation results of sparse coding, archetypal analysis and our method. The x-axis is the reconstruction error. The y-axis is the percentage of data whose estimation errors are less equal than the error specified by the x-axis. The right figure shows the average limb length of the estimated 3D poses of the three methods and ground-truth. The limb lengths of our method are much closer to the ground-truth which shows that the model better respects the anthropomorphic constraints.}
\label{fig:pose_estimation}
\vspace{-1em}
\end{figure}

We represent the 3D poses by activated simplices. We first learn a set of activated simplices $\mathbb{S}=\{\mathcal{S}_1,\cdots,\mathcal{S}_K\}$ from the training 3D poses. Then for a 2D pose $O$,  we optimize for the simplex $\mathcal{S}_* \in \mathbb{S}$ and the simplex coefficients $\beta_*$ by minimizing the following:
\begin{equation}
\begin{aligned}
\mathcal{S}_*, \beta_* = & \underset{\mathcal{S} \in \mathbb{S}, \beta}{\text{argmin}} \underset{ }{}
& & { \left\| M {\mathcal{S}} {\beta} - O \right\|_2^2 } \\
& \text{s.t.}
& & {\beta} \geq 0, \quad \|\beta\|_1=1
\end{aligned}
\label{eq:2D_recon}
\end{equation}
This minimization can be done naively since the number of simplices is small. Then we reconstruct its 3D pose $P$ by $\hat{P}=\mathcal{S}_* \beta_*$. The reconstruction error is the average Euclidean distance between each joint of $P$ and $\hat{P}$.

We compare our method with the classical sparse coding \cite{Mairal:2009um} and archetypal analysis \cite{cutler1994archetypal} methods. For these two methods, we first learn a set of bases or archetypes $X$ on the training data and then for each test data, we optimize the basis coefficients $\beta_*$ by minimizing the error between the 3D pose projection and the 2D pose:

\begin{equation}
\begin{aligned}
\beta_* = & \underset{\mathcal{\beta}}{\text{argmin}}
{ \left\| M {X} {\beta} - O \right\|_2^2 .} \\
\end{aligned}
\end{equation}
For Archetypal method we constrain $\beta$ is non-negative and sum to one. The estimated 3D pose is $\hat{P}=X \beta_*$.

Figure \ref{fig:pose_estimation} (left) shows the human pose estimation results. Our method achieves better performance than the other two alternatives. There could be two reasons accounting for this: (1) compared with sparse coding, the activated simplices constrain the combinations of bases to combinations that occur in reality, whereas the restrictions in sparse coding methods are looser. Figure \ref{fig:pose_estimation} (right) shows that the estimated limb lengths of our method are close to the ground-truth while those of the sparse coding and archetypes are considerably different from the ground-truth. The experimental results show that the sparse representation cannot implicitly enforce the limb length constraint; (2) compared with archetypal analysis method, the activated simplices explore the local structures including the interior of the manifold which makes the method more accurate. In contrast, archetype can only learn the convex hull of the manifold and fail to explore the interior.


\section{3D Human Pose Synthesis}
\label{sec:synthesize}
Our method also allows to synthesize realistic data. For each simplex $\mathcal{S}_i$, we learn a Dirichlet distribution \cite{minka2000estimating} from the coefficients $\beta_i^j, j=1,\cdots,m$ of the $m$ data that are projected to the simplex.
To generate a data, we first sample a simplex and then sample the coefficients $\beta$ from the Dirichlet distribution of that simplex. We output the corresponding combination of bases as the synthesized data.

Dirichlet distribution is often used as prior distributions in Bayesian statistics. The probability density function is defined as:
$P(\beta)=\frac{1}{B(\alpha)}{\prod_{i=1}^{m}{\beta_i^{\alpha_i-1}}}$, where $\beta > 0$ and $\| \beta \|_1=1$. $B(\alpha)$ is a normalization factor. We learn the parameters $\alpha$ by Maximum Likelihood Estimation.

\begin{figure}
\centering
\includegraphics[scale =0.25]{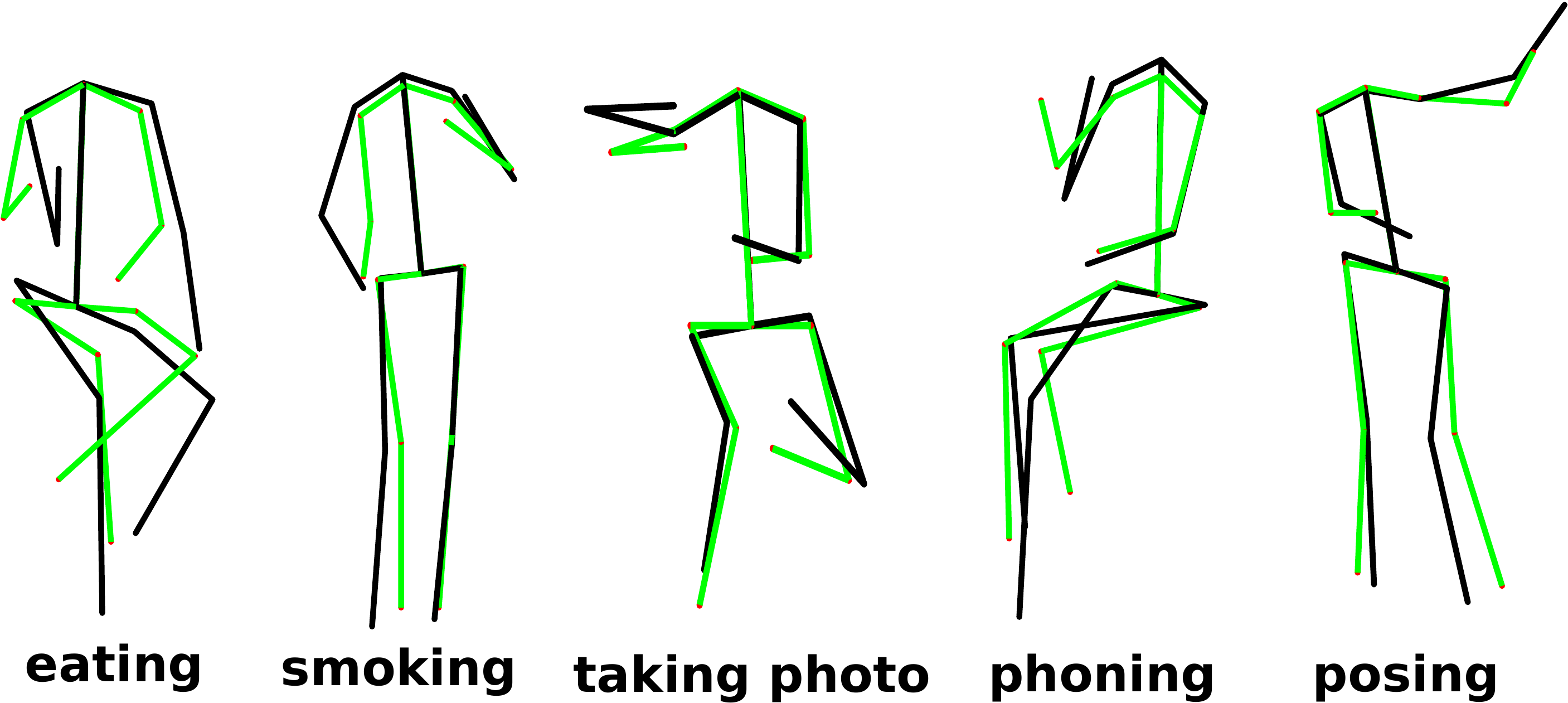}
\caption{Synthesized human poses (in green) are overlaid with their nearest neighbours (in black) in the training dataset. The actions of the nearest neighbours are shown in the texts below the poses. We can see that the synthesized poses differ from the training data but they are still realistic which shows the simplices' generalization properties.}
\label{fig:synthesis}
\end{figure}

\begin{figure}
\centering
\includegraphics[scale =0.3]{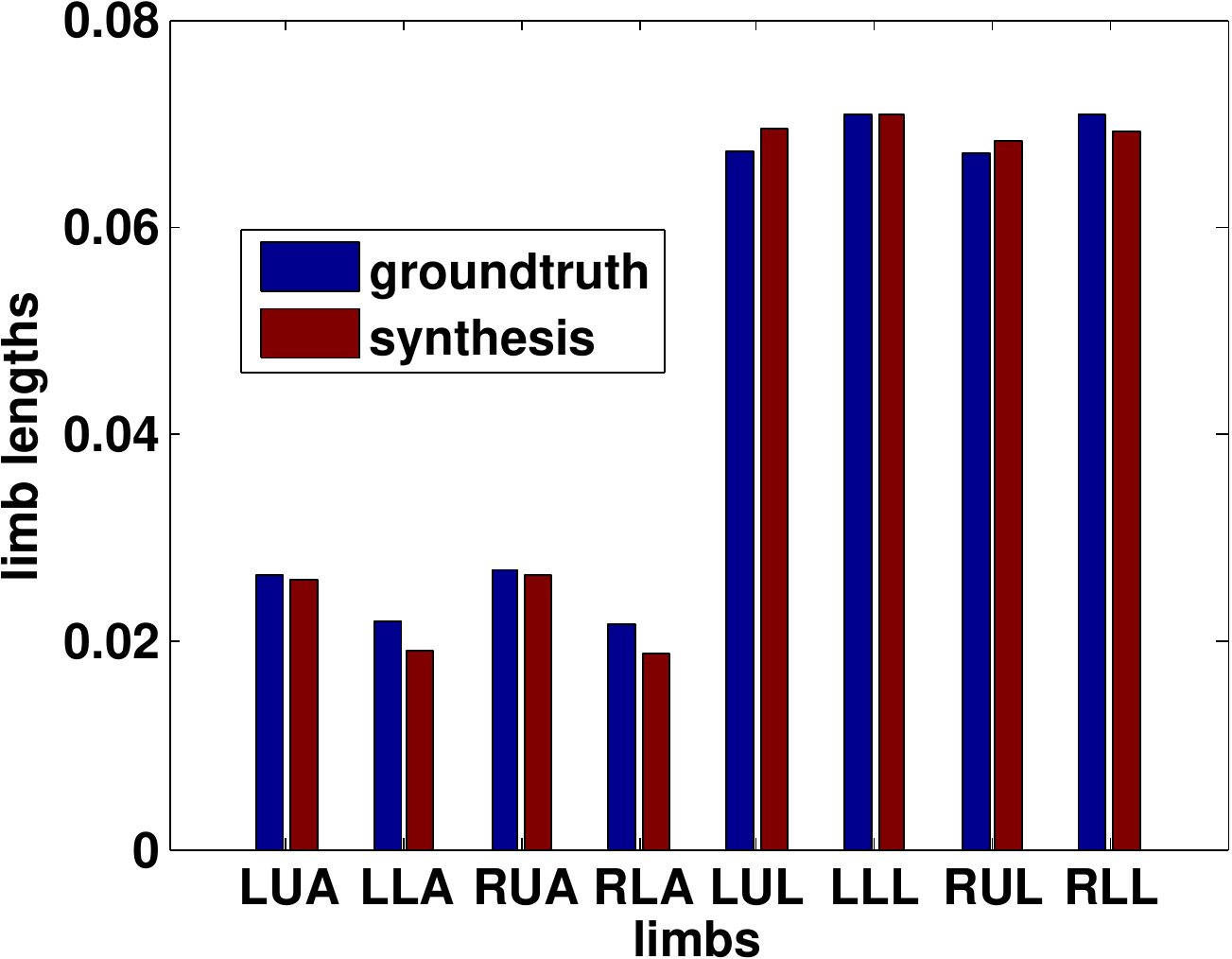}
\includegraphics[scale =0.3]{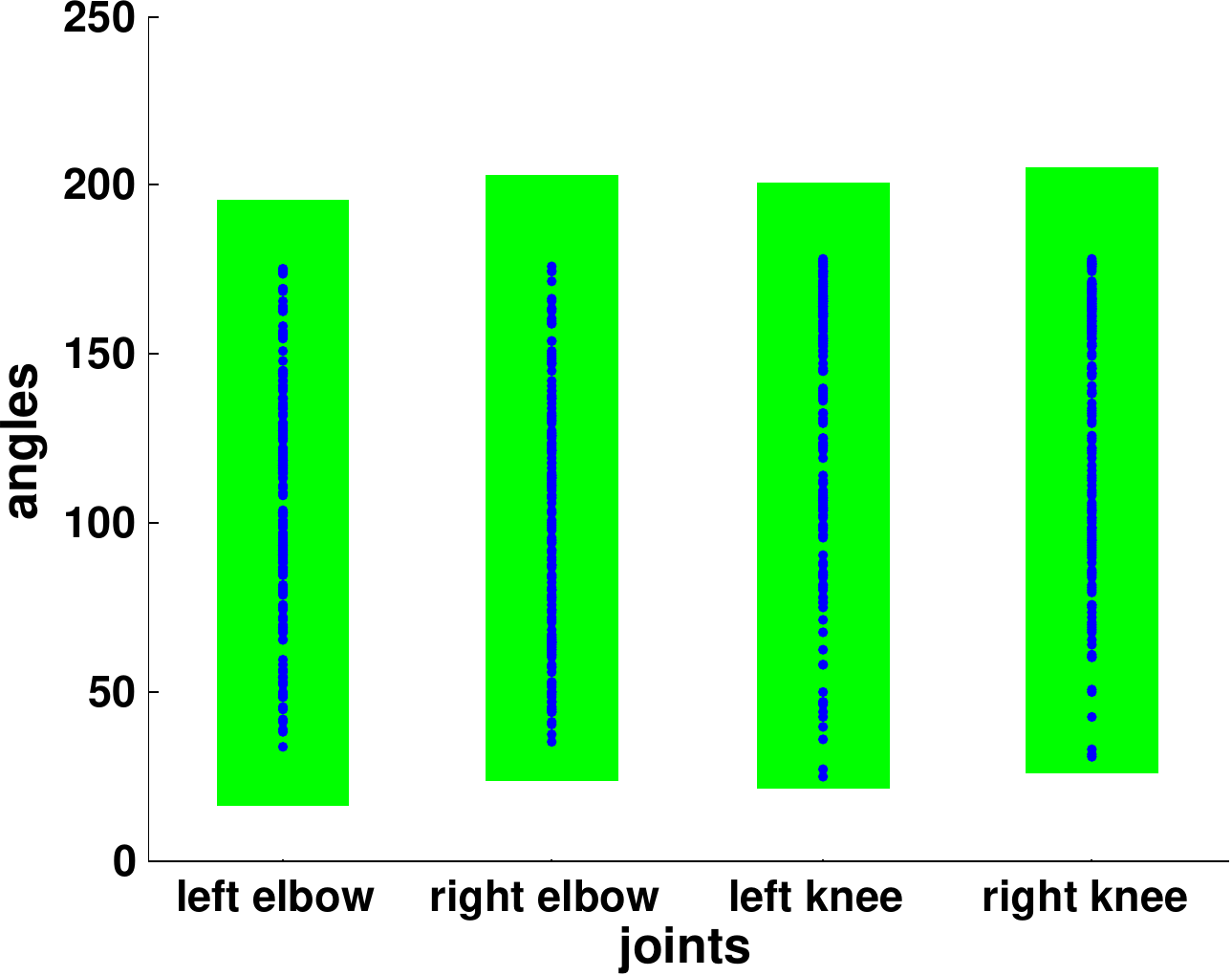}
\caption{Left figure shows the limb lengths of the synthesized poses and of the training data (groundtruth). The right figure shows the range of bending angles of four body joints in the training data (green bars) and the bending angles of the synthesized data (blue dots). We can see that the bending angles of the synthesized data are within the limit of those of the training data. }
\label{fig:synthesize_boundary}
\vspace{-1em}
\end{figure}

We conduct experiments on the H3.6M human pose dataset. We select $55,000$ poses from $11$ actions and learn a single set of simplices. We set the number of bases to be $100$. The algorithm ends up with $83$ simplicies. We learn Dirichlet distributions on each simplex. We display some typical synthesized poses in Figure \ref{fig:synthesis}. We can see that they are divergent and realistic. In addition, Figure \ref{fig:synthesize_boundary} shows that the synthesized poses also respect well the limb length and bending angle constraints.

\section{Conclusion}
\label{sec:conclusion}

We propose a method for representing data using a mixture of activated simplices. The Activated Simplices representation allows accurate reconstruction by preventing combinations of bases that do not occur in the data.

\textbf{Acknowledgement: The project is supported by US Army Research Office ARO Proposal 62250-CS and NSF STC award CCF-1231216 and Center for Minds, Brains and Machines (CBMM)}

\clearpage
\textbf{\Huge{Appendix starts here}}

\section{Introduction}

In the appendix we provide more details on the following:
\begin{enumerate}[(i)]
\item The relationship between Activated Simplices and traditional sparse coding.
\item Alternatives to standard normalization for projection onto the sphere.
\item Triangulating Manifolds.
\item Extremes in data and a comparison with Archetypal Analysis.
\end{enumerate}

\section{Activated Simplices and Sparse Coding}

In Section 3 of the Main Paper we discussed how a mixture of simplices model can be learned by constructing a convex hull inside the sphere whose boundary facets are close to the data. This will remind readers of  traditional sparse coding \cite{olshausen1996emergence}, since the  projections of the training data on this hull are sparse, i.e., they involve relatively few of the bases. We now explore this connection in more detail.

Our model requires normalized data $Y = \{y^{(1)},\ldots, y^{(N)}\}$ and it constructs a convex hull of $p$ bases $X = \{x_1, \ldots, x_p\}$  such that boundary facets are close to the data. It constructs this hull by the following minimization
\begin{equation}
\label{eq:newformulation}
\min_{X, \beta} \frac{1}{N} \sum _{j=1}^N ||y^{(j)} -X\beta^{(j)}||^2
\end{equation}
subject to $\beta^{(j)} \ge 0$, $\|\beta^{(j)}\|_1 =1$, for $j = 1, \ldots, N$, and $\|x_i\|_2 \le 1$ for $ i = 1, \ldots, p$.\\
The activated simplices correspond to co-activated bases.

\begin{figure}
\centering
\includegraphics[height=1.4in]{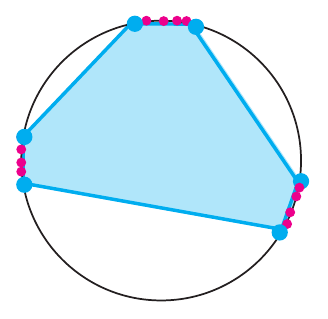}
\caption{A simple picture showing data on the circle (red dots) and a convex hull fitted to the data using 6 bases (blue dots).}
\label{fig:simple_data_on_sphere}
\vspace{-1em}
\end{figure}

Figure \ref{fig:simple_data_on_sphere} shows a very simple situation where a convex hull is constructed from data on the circle using 6 bases. The activated simplices are the three boundary segments closest to the data.

In the usual formulation of sparse coding, a dictionary of $p$ bases $X =\{x_1, \ldots, x_p\}$ is constructed to represent data $Y = \{y^{(1)},\ldots, y^{(N)}\}$. The dictionary is constructed by the following minimization
\begin{equation}
\min_{X,\beta} \frac{1}{N} \sum _{i=1}^N \left(\frac{1}{2}||y^{(j)} -X\beta^{(j)}||_2^2 + \lambda ||\beta^{(j)}||_1\right)
\label{eq:standardsparsecoding}
\end{equation}
subject to $\|x_i\|_2 \le 1$ for $ i = 1, \ldots, p$. Here, $\lambda$ is a non-negative penalty parameter.

Notice some differences between (\ref{eq:newformulation}) and (\ref{eq:standardsparsecoding}). In (\ref{eq:newformulation}) the coefficients $\beta^{(j)}$ are required to be non-negative, but there is no such constraint in (\ref{eq:standardsparsecoding}). In (\ref{eq:newformulation}) the coefficients $\beta^{(j)}$ are constrained to have $\ell_1$ norm 1, whereas in (\ref{eq:standardsparsecoding}) the $\ell_1$ norms of the coefficients are penalized through the $ \lambda ||\beta^{(j)}||_1$ term. But both formulations have a similar quadratic data reconstruction term  $||y^{(j)} -X\beta^{(j)}||_2^2$.

It is possible to interpret this sparse coding in geometrical terms \cite{Donoho:2005we, Huggins}, and
it is likely that this geometrical understanding was behind Tibshorani's original formulation of the lasso \cite{tibshirani1996regression}.

To begin, we describe the geometry of a single penalized regression. In this situation the bases $X$ are known, and we are interested in the coefficients $\beta$ for a single data point $y$. This corresponds to the minimization \cite{tibshirani1996regression, Osborne:2000ha}
\begin{equation}
\label{eq:penalizedregression}
\min_{\beta}\left( ||y -X\beta||^2 +2 \lambda ||\beta||_1\right)
\end{equation}
In this regression the penalty parameter $\lambda$ and the data point $y$ control the $\ell_1$ norm of the coefficients, that is, they determines a {\em radius} $r = r(y,\lambda)$ such that the coefficients $\beta$ solve
\begin{equation}
\label{eq:penalizedregressionexpanded}
\min_{\beta} ||y -X\beta||_2
\end{equation}
subject to $ \|\beta\|_1 \le r$. As the penalty parameter $\lambda$ increases, the radius $r$ decreases.

\begin{figure}
\centering
\includegraphics[height=1.4in]{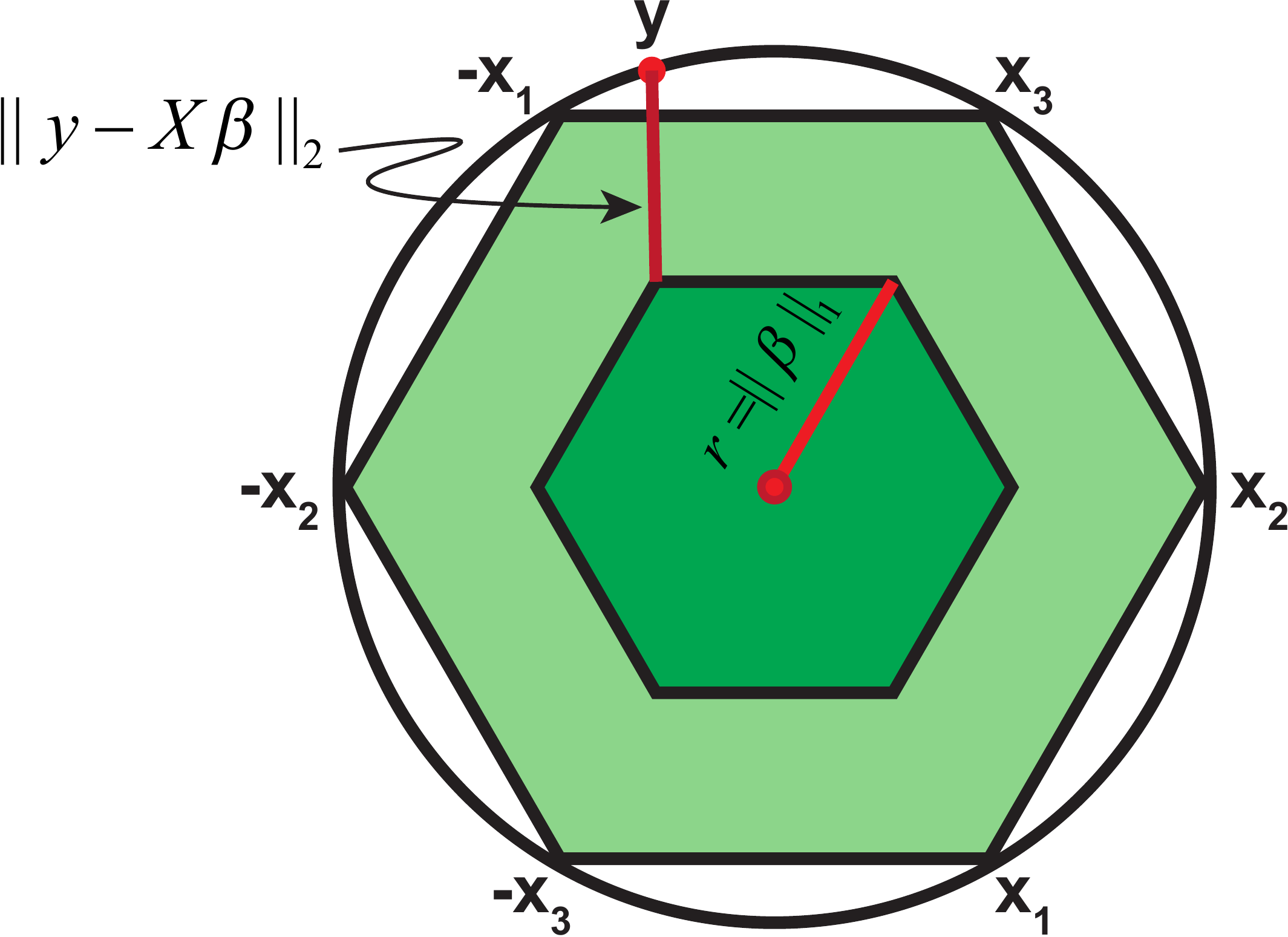}
\caption{The convex hull of bases and negative bases scaled by $r$.}
\label{fig:basesnegbases}
\vspace{-1em}
\end{figure}

We can interpret this minimization in terms of convex geometry. The set of $X\beta$ such that $\|\beta\|_1 \le r$ is the convex hull of the bases $X = \{x_1,\ldots, x_p\}$ and the negative bases $-X = \{-x_1,\ldots, -x_p\}$, scaled by a factor of $r$. Figure \ref{fig:basesnegbases} shows bases, and negative bases, and their convex hull, scaled by $r$.

For the regression of a single point, the penalty $\lambda$ determines a convex hull of radius $r$, and then the regression coefficients $\beta$ correspond to the closest point on the boundary of the scaled convex hull. See Equation \ref{eq:penalizedregressionexpanded}.

\begin{figure}
\centering
\includegraphics[height=1.4in]{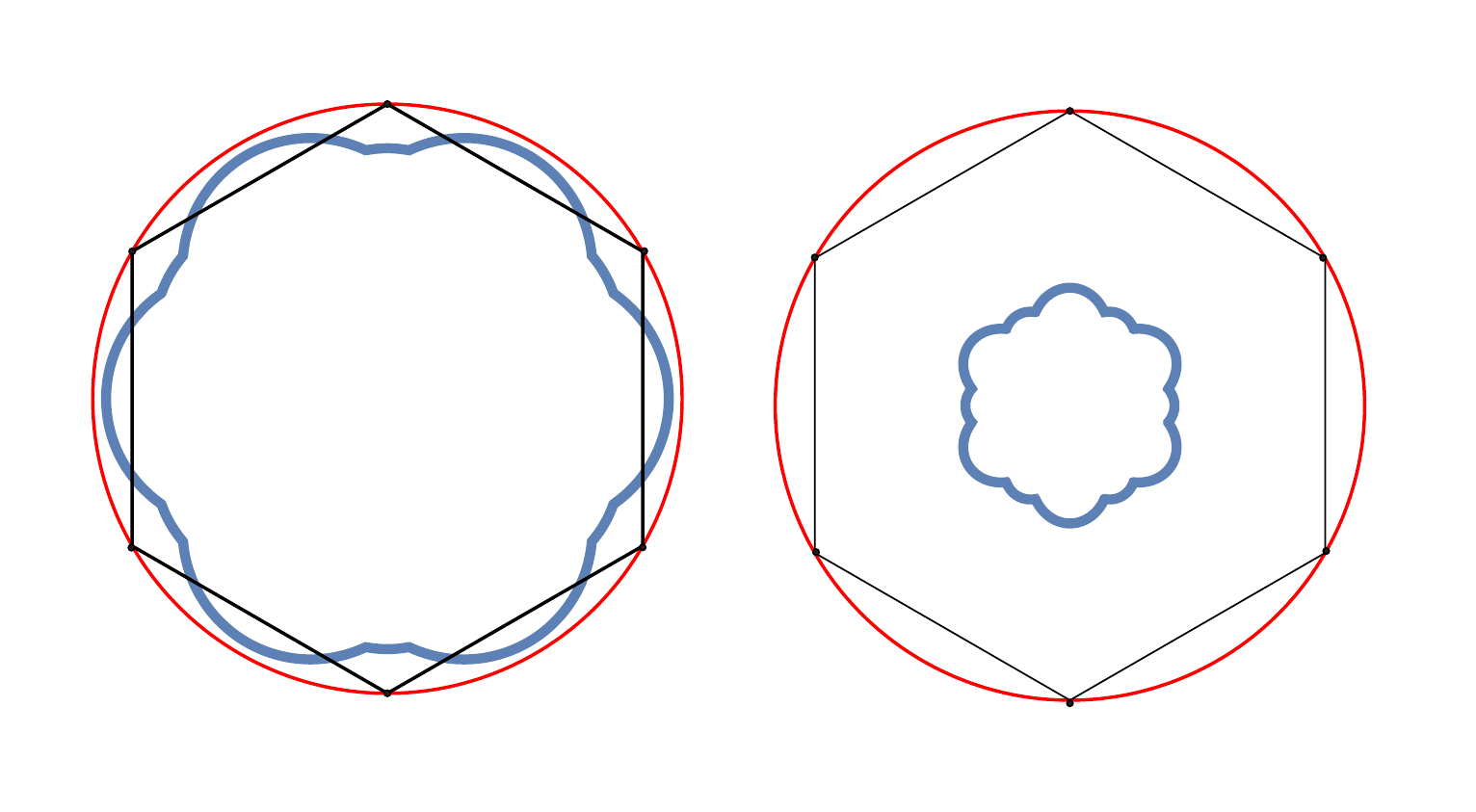}
\caption{The radius $r$ (blue curve) in penalized regression depends on $\lambda$ and $y$. On left $\lambda =0.15$, on right $\lambda =0.6$.}
\label{fig:variation_in_radius}
\vspace{-1em}
\end{figure}

The geometry is more complicated when more than one point is regressed on $X$, as the radius $r$ depends on the penalty $\lambda$ and on the data point $y$. Figure \ref{fig:variation_in_radius} shows that $r$ varies with $y$. In this figure points on the circle are regressed on 3 bases with $\lambda = 0.15$ and $\lambda =0.6$. The blue curve shows how $r(y)$ varies as $y$ moves along the circle.

This gives us a geometrical understanding (albeit a somewhat complicated one) of the learning process  in the usual formulation of sparse coding. The method positions the convex hull of the bases and negative bases, so that boundary facets of some scalings of this hull are close to the data. The complication here is that the scalings are not uniform, they depend on the $y^{(j)}$.

Our method (\ref{eq:newformulation}) makes two major changes to the sparse coding formulation. First, the radius $r$ is set uniformly to 1 for all data points $y^{(j)}$, thus we can think of the data as projecting on the same convex hull. Second, the coefficients $\beta^{(j)}$ are restricted to be non-negative, thus we consider the convex hull of the bases, rather than the hull of the bases and negative bases.

\begin{figure}
\centering
\includegraphics[height=1.4in]{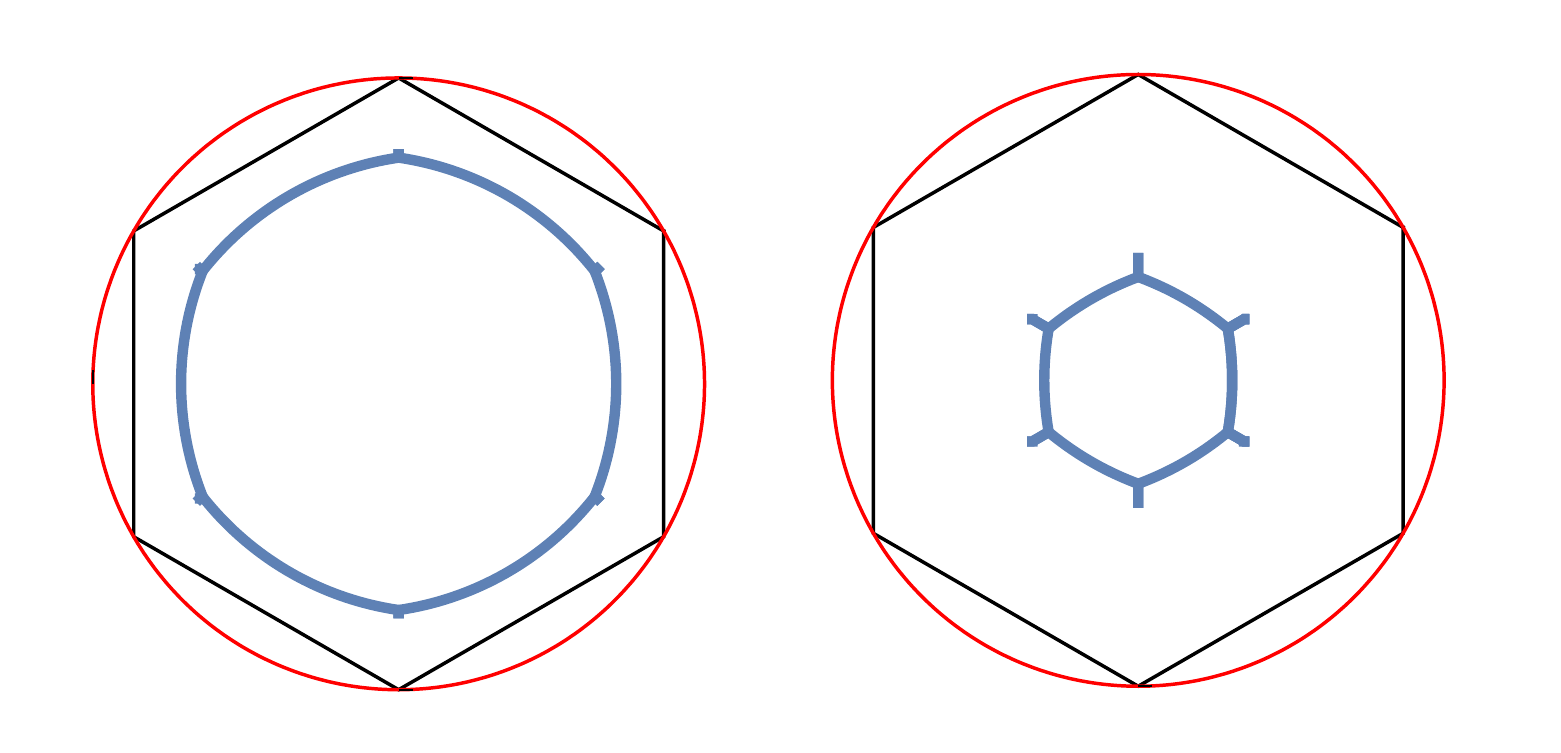}
\caption{The projection (blue curve) of the circle learned using 3 bases and their negatives. Notice the bulging. The left panel uses  $\lambda = 0.25$ and the right uses $\lambda =0.6$. }
\label{fig:projection_plot}
\vspace{-1em}
\end{figure}

We might have learned co-activations from the original sparse coding formulation, but the modifications we've made improve performance as well as interpretability. The uniform radius requirement $r=1$  encourages the fitting of low-dimensional facets to the data. This encouragement is weaker in the usual sparse coding. Figure \ref{fig:projection_plot} shows the projection (blue curves) of the circle learned using the usual sparse coding with three bases and their negatives. Notice that the projections are no longer polygonal. This fitting of a bulging polytope lessens the incentive to fit low dimensional facets close to the data. The left panel in Figure  \ref{fig:projection_plot} uses $\lambda = 0.25$ and the right panel used $\lambda = 0.6$.

The second modification restricts the coefficients $\beta^{(j)}$ to be non-negative.  Without this restriction, a basis might be used with a positive coefficient in some combination to approximate data in some region of the data manifold and it might be used with a negative coefficient to represent data somewhere else, far off on the data manifold. The positive restriction means that bases are used in a more focused way  in reconstruction and learning.

\subsection{Penalized Activated Simplices}

We presented a penalized version of  our activated simplices formulation  in Section 3.3 of the Main Paper. We might have applied an $\ell_1$ penalty as in the usual sparse coding (\ref{eq:standardsparsecoding}), but we prefer to control the geometry uniformly  and explicitly through $r$. The penalized formulation with penalty parameter  $r \le 1$ is
\begin{equation}
\label{eq:spenalizedactivatedsimplices}
\min_{X, \beta} \frac{1}{N} \sum _{j=1}^N \|y^{(j)} -X\beta^{(j)}\|^2
\end{equation}
subject to $\beta^{(j)} \ge 0$, $\|\beta^{(j)}\|_1 =r$, for $j = 1, \ldots, N$,
 and $\|x_i\|_2 \le 1$ for $ i = 1, \ldots, p$.

\begin{figure}
\centering
\includegraphics[height=1.4in]{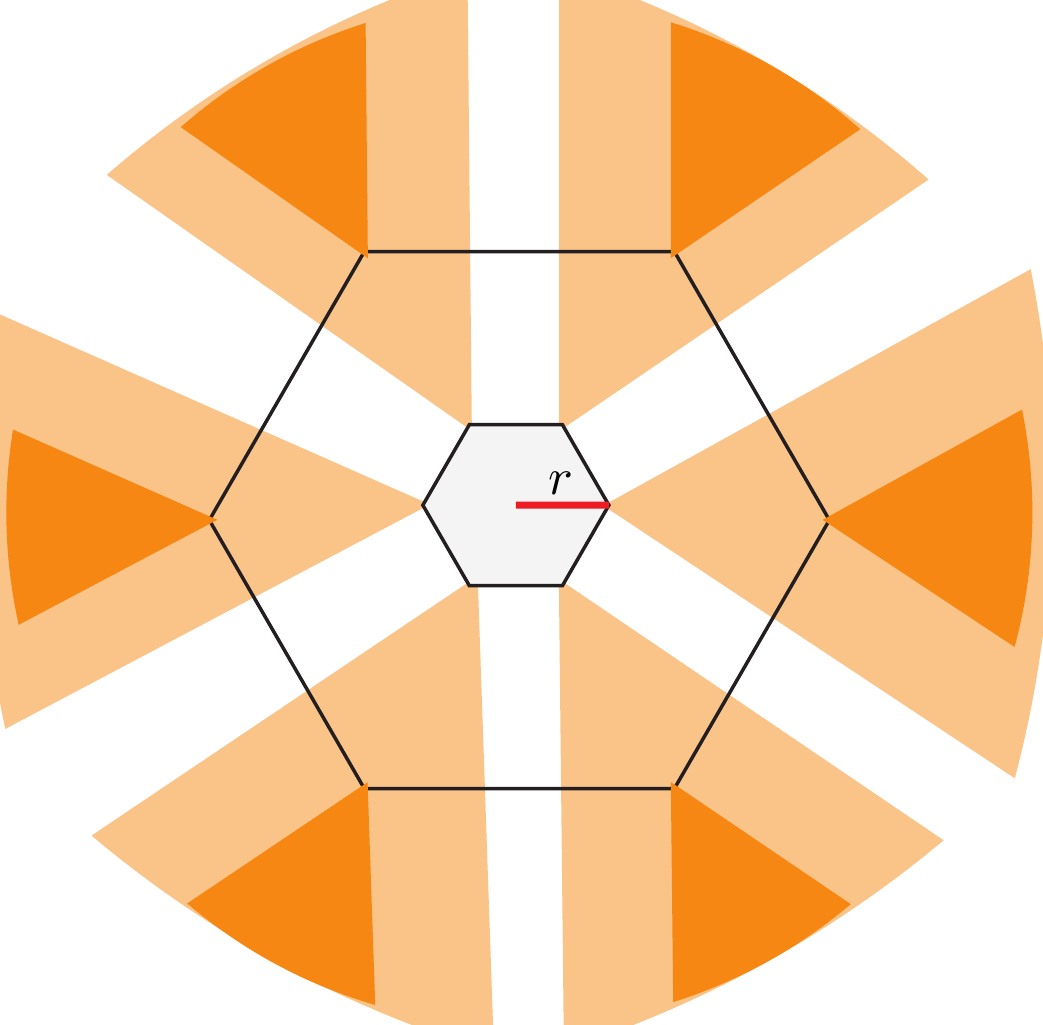}
\caption{Decreasing $r$ encourages sparsity. Points on the orange cones project to $0$-simplices.}
\label{fig:sparsity_shrinkage}
\vspace{-1em}
\end{figure}

This can be understood geometrically as learning a convex hull of radius $r$, such that boundary facets are close to the data. Figure \ref{fig:sparsity_shrinkage} shows how decreasing $r$ encourages sparsity. In the figure, points in the orange regions project onto the vertices, and as $r$ decreases more of space is covered by these orange regions.

This penalization will necessarily worsen reconstruction, but it is useful if one is interested in dimension reduction. In our experiments, which focused on reconstruction from real low dimensional data sources, we found that $r=1$ was best. See Figure 9 and 11 of the Main Paper where we experiment with the number of bases $p$ and the radius $r$ for reconstruction of poses and digits. We experiment with $r$ in reconstructions of the torus from synthetic data in Section \ref{sec:alternative_projections}.

\section{Alternative Projections}
\label{sec:alternative_projections}

The Activated Simplices method requires that training data lie on the sphere. This is usually accomplished by directly normalizing the data, or by centering and then normalizing.

The standard normalization is the transformation
\begin{equation}
y^{(j)} \mapsto \frac{y^{(j)}}{\|y^{(j)}\|_2}
\label{eq:normalization}
\end{equation}
and centering and normalizing is the transformation
\begin{equation}
y^{(j)} \mapsto \frac{y^{(j)}- \overline{y}}{\|y^{(j)} -\overline{y}\|_2}
\label{centering_and_normalization}
\end{equation}
where $\overline y$ is the data mean $\frac{1}{N} \sum_{j=1}^N y^{(j)}$.

For many signals little information is lost in replacing data by its normalization. Indeed this normalization is a common first step in many analyses, for example, images are often contrast normalized. In the experiments in this paper the data was centered and normalized.

\begin{figure}
\centering
\includegraphics[height=1.4in]{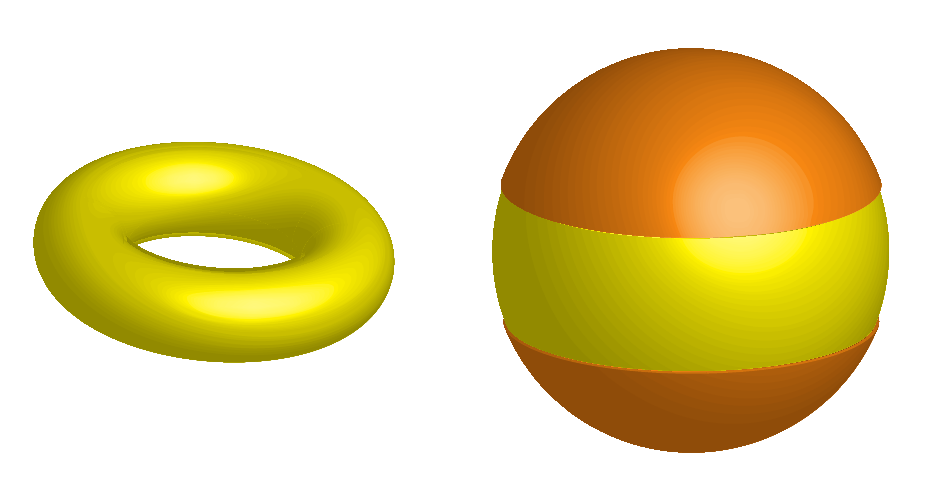}
\caption{The torus is flattened into a band when normalized.}
\label{fig:torus_sphere}
\vspace{-1em}
\end{figure}

However some structure may collapse under direct normalization.  Consider, for example, the usual 2 dimensional torus. Figure \ref{fig:torus_sphere} shows the torus and its normalized image in the sphere  --- the normalization flattens the torus into a band around the equator and the inner cavity is lost.

\begin{figure}
\centering
\includegraphics[height=1.4in]{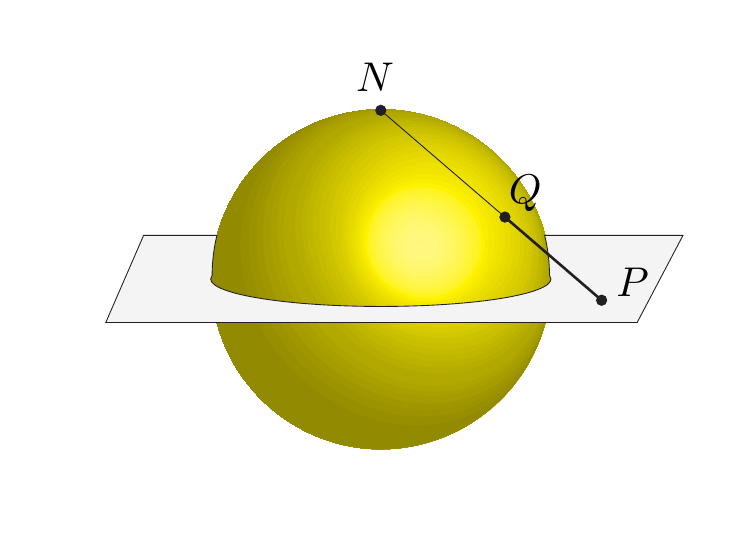}
\caption{Stereographic projection from the equatorial plane into the sphere maps $P$ to $Q$.}
\label{fig:stereographic_projection}
\vspace{-1em}
\end{figure}

To avoid flattenings such as this we use {\em stereographic projection} to map data in $\R^d$ into the $d$ dimensional sphere $\sphere^d$ in $\R^{d+1}$. This is a standard technique in map making where our world (the 2-dimensional sphere) is mapped onto a flat sheet ($\R^2$). Figure \ref{fig:stereographic_projection} shows stereographic projection from the equatorial plane onto the sphere, through the north pole; the point $P$ maps to the point $Q$. The point $Q$ is the point on the line through $N$ and $P$ that lies on the sphere.

Stereographic projection maps $\R^d$ onto the $d$-sphere with the north pole removed (the points at infinity map to the north pole.) It is an invertible conformal map (it preserves angles between curves). The map is
\begin{equation}
 \pi: P\mapsto  Q =N + \frac{2}{1 + \|P\|^2} (P-N)
\label{eq:stereographic_projection}
\end{equation}
and the inverse is
\begin{equation}
  \pi^{-1}:Q\mapsto  P=N + \frac{1}{1 - Q_{d+1}} (Q-N)
\label{eq:inverse_stereographic_projection}
\end{equation}
where $Q_{d+1}$ is the $d+1^{\text{st}}$ coordinate of $Q$.

In some situations the entire data analysis can be carried out in $\sphere^d$ rather than in $\R^d$. For a classification task, one might map the data to the sphere with stereographic projection, learn a simplicial structure, and carry out classification tasks there. For synthesis one might map the synthesized points back to $\R^d$ with inverse stereographic projection.

\begin{figure}
\centering
\includegraphics[height=1.4in]{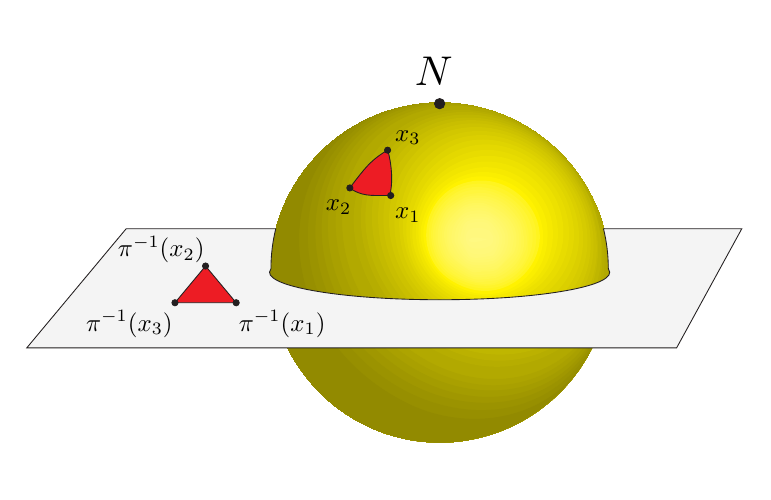}
\caption{A simplex in the plane and the corresponding curved simplex on the sphere.}
\label{fig:simplex_to_simplex}
\vspace{-1em}
\end{figure}

But it can also be useful to map the simplicial structure back to $\R^d$. This is simply a matter of keeping track of activations. The simplex in $\R^d$ interpolating  $\pi^{-1}(x_{i_0}), \ldots , \pi^{-1}(x_{i_k})$ maps to the curved simplex in $\sphere^d$ with vertices $x_{i_1},\ldots, x_{i_k}$, which lies over the flat activated simplex with vertices $x_{i_1},\ldots, x_{i_k}$ on the convex hull! Figure \ref{fig:simplex_to_simplex} illustrates the relationship between a simplex in the plane and a curved simplex on the sphere.

This relationship between a simplicial structure in Euclidean space and a convex structure in the sphere will remind some readers of \cite{Brown:1979dj}, where it is revealed that the Delaunay triangulation for some data can be constructed by mapping the data into the sphere with stereographic projection, finding the convex hull of the projected points, and mapping the faces of this hull back to Euclidean space. Our method approximates the convex hull of the projected points and builds a simplicial approximation to the data in Euclidean space by mapping some facets  of this hull back to Euclidean space. It is also interesting to compare with \cite{Edelsbrunner:1986wt} which shows that the Delaunay triangulation can be found from a convex hull after mapping the data into a paraboloid. The common theme is that mapping data into the boundary hyper-surface of a convex region allows useful triangulations to be constructed by convex approximation.

We'll say a little more about Delaunay triangulations in Section \ref{sec:triangulating_manifolds}.

\begin{figure}
\centering
\includegraphics[height=1.8in]{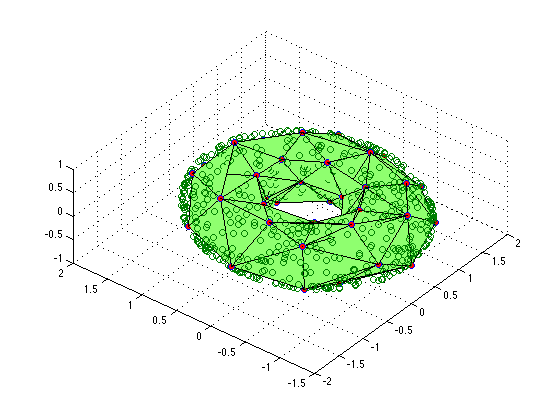}
\caption{torus using $N=1000, p =60, r=1$}
\label{fig:torus_1}
\vspace{-1em}
\end{figure}

\begin{figure}
\centering
\includegraphics[height=1.8in]{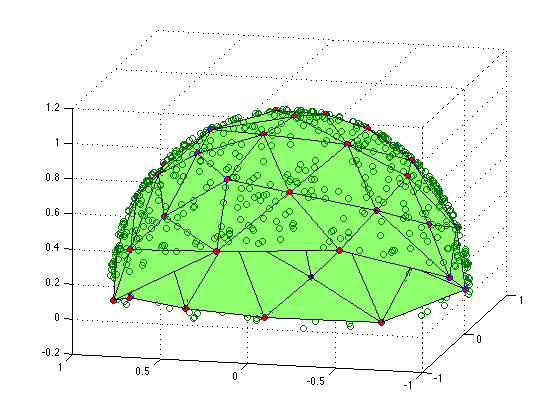}
\caption{hemisphere using $N=1000, p =50, r=1$}
\label{fig:hemisphere_1}
\vspace{-1em}
\end{figure}

\begin{figure}
\centering
\includegraphics[height=2in]{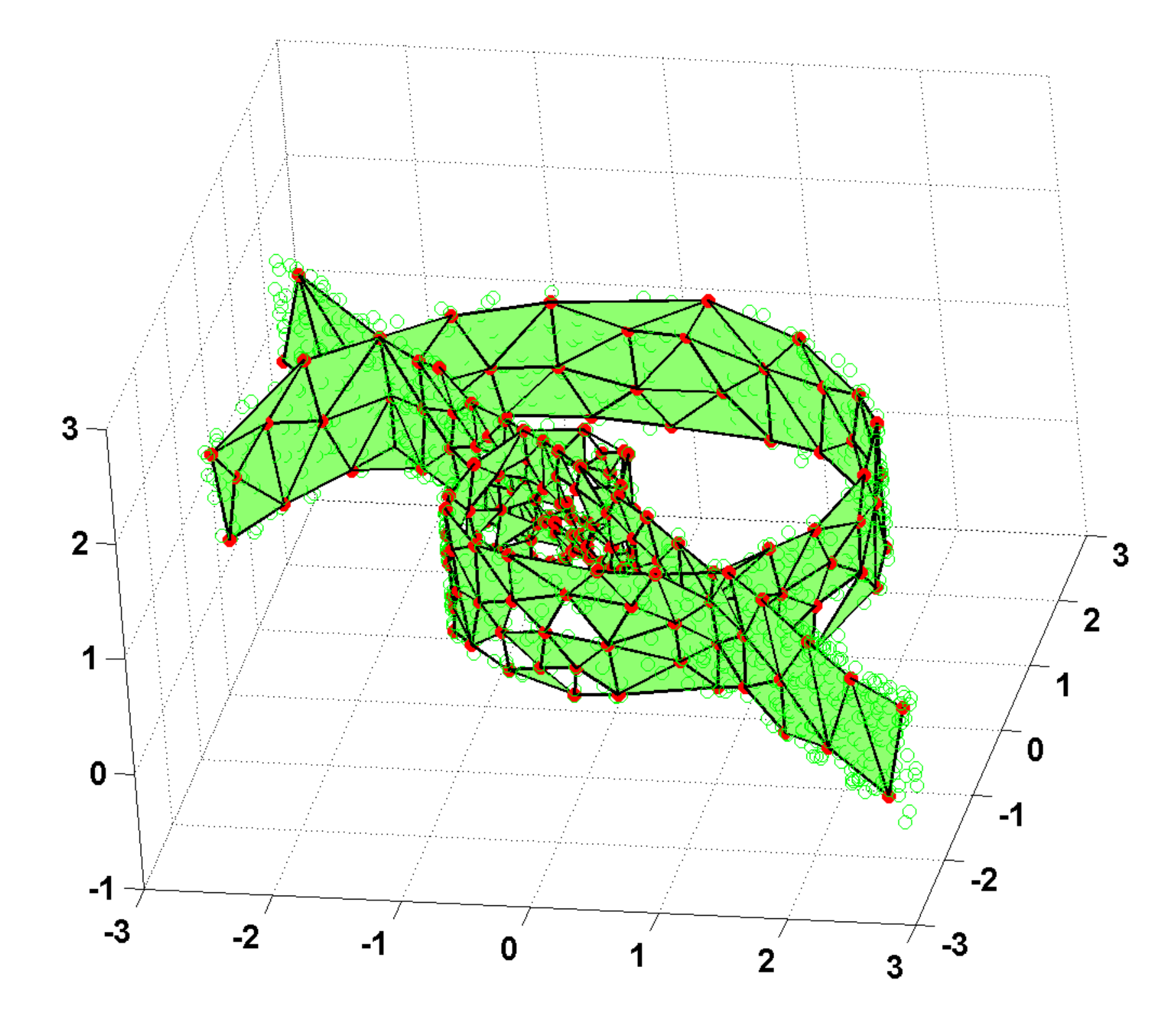}
\includegraphics[height=2in]{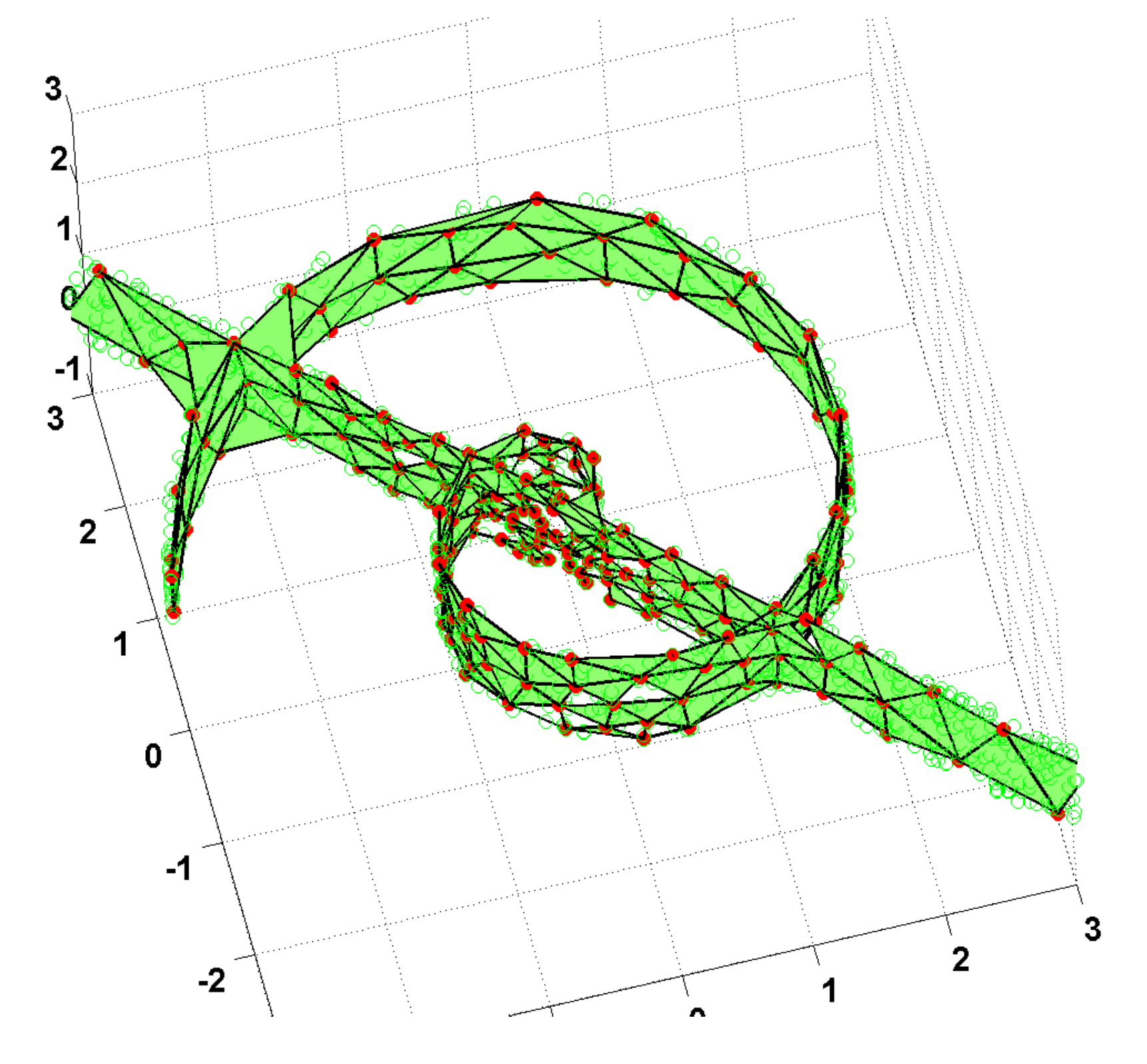}
\caption{Swissroll and plane using $N=2000, p =100, r=1$ viewed from two directions.}
\label{fig:swiss_roll_and_plane}
\vspace{-1em}
\end{figure}

\subsection{Low Dimensional Examples.}
\label{sec:low_dimensional}

We now discuss some  experiments with some data from some low dimensional manifolds. We sample data from some 2 dimensional manifolds embedded in $\R^3$; we map the data into the 3-dimensional sphere $\sphere^3$ using stereographic projection and build a simplicial structure; we map the simplicial structure back into $\R^3$ and show some plots. No pruning was used here.

Figure \ref{fig:torus_1} shows a structure learned form $N=1000$ data points sampled from the torus, with $p=60$ bases using $r=1$ (unpenalized), and with no pruning. Some 3-dimensional simplices were activated but only their 2-dimensional boundaries are shown here.

Figure \ref{fig:hemisphere_1} shows a simplicial structure learned from 1000 points sampled from the hemisphere, using 50 bases and $r=1$. No simplices of dimension more than 2 were activated and there are no holes in the structure. This is an excellent reconstruction of the hemisphere.

Figure \ref{fig:swiss_roll_and_plane} shows a structure learned from 2000 points sampled on the union of a ribbon and a planar segment, using $100$ bases and $r=1$. A small number of 3-dimensional simplices were learned.

Figure \ref{fig:torus_reconstructions} shows simplicial structures learned from  the torus, using 1000 samples, and for several values of $p$ and $r$. Notice that as $r$ gets smaller the activations become more sparse, and for small values of $r$ the structure is a wireframe. A heavily penalized structure constructed using 5 bases and $r=0.4$ and $r=0.1$ captures the medial axis of the torus.


\section{Triangulating Manifolds}
\label{sec:triangulating_manifolds}

In the main paper we focused on  Activated Simplices as a method for data reconstruction, regularization and synthesis. We now discuss Activated Simplices as a tool for manifold reconstruction. This hasn't been a priority in this work, but it is a natural extension. The illustrations in Section \ref{sec:low_dimensional} show the potential for reconstructing low dimensional manifolds.  Figure (\ref{fig:swiss_roll_and_plane}) shows a reconstruction of a structure with intersecting components.

\begin{figure}
\centering
\includegraphics[height=1.3in]{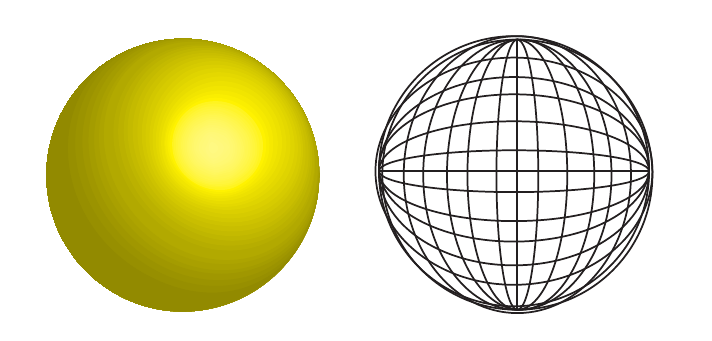}
\caption{A sphere and a wireframe. The wireframe is good enough for data regularization, but it is a poor reconstruction of the manifold.}
\label{fig:sphere_wireframe}
\vspace{-1em}
\end{figure}

It is useful to distinguish the task to building a structure for data reconstruction and regularization from that of manifold reconstruction. Figure \ref{fig:sphere_wireframe} shows a sphere and a wireframe. A fine wireframe is a good enough structure to regularize data from the sphere but it is a poor reconstruction of the manifold. The goal in manifold reconstruction is usually to construct a smooth manifold, or a union of a small number of  smooth manifolds, that approximate a data source.

The pruning process in Section 4 of the Main Paper is aimed at economical reconstruction, but it may be  inappropriate for manifold reconstruction where  smoothness is a priority. The plots in Section \ref{sec:low_dimensional} show unpruned structures.

\begin{figure}
\centering
\includegraphics[height=1.5in]{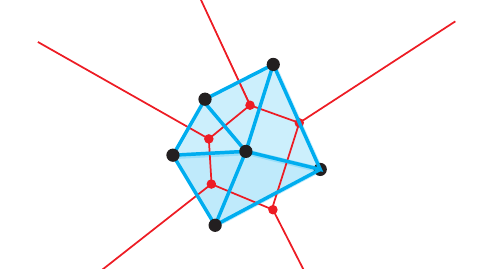}
\caption{Some points (black dots), the Voronoi boundaries (red) and the Delaunay triangulation (blue)}
\label{fig:voronoidelaunay}
\vspace{-1em}
\end{figure}

\begin{figure}
\centering
\includegraphics[height=1.3in]{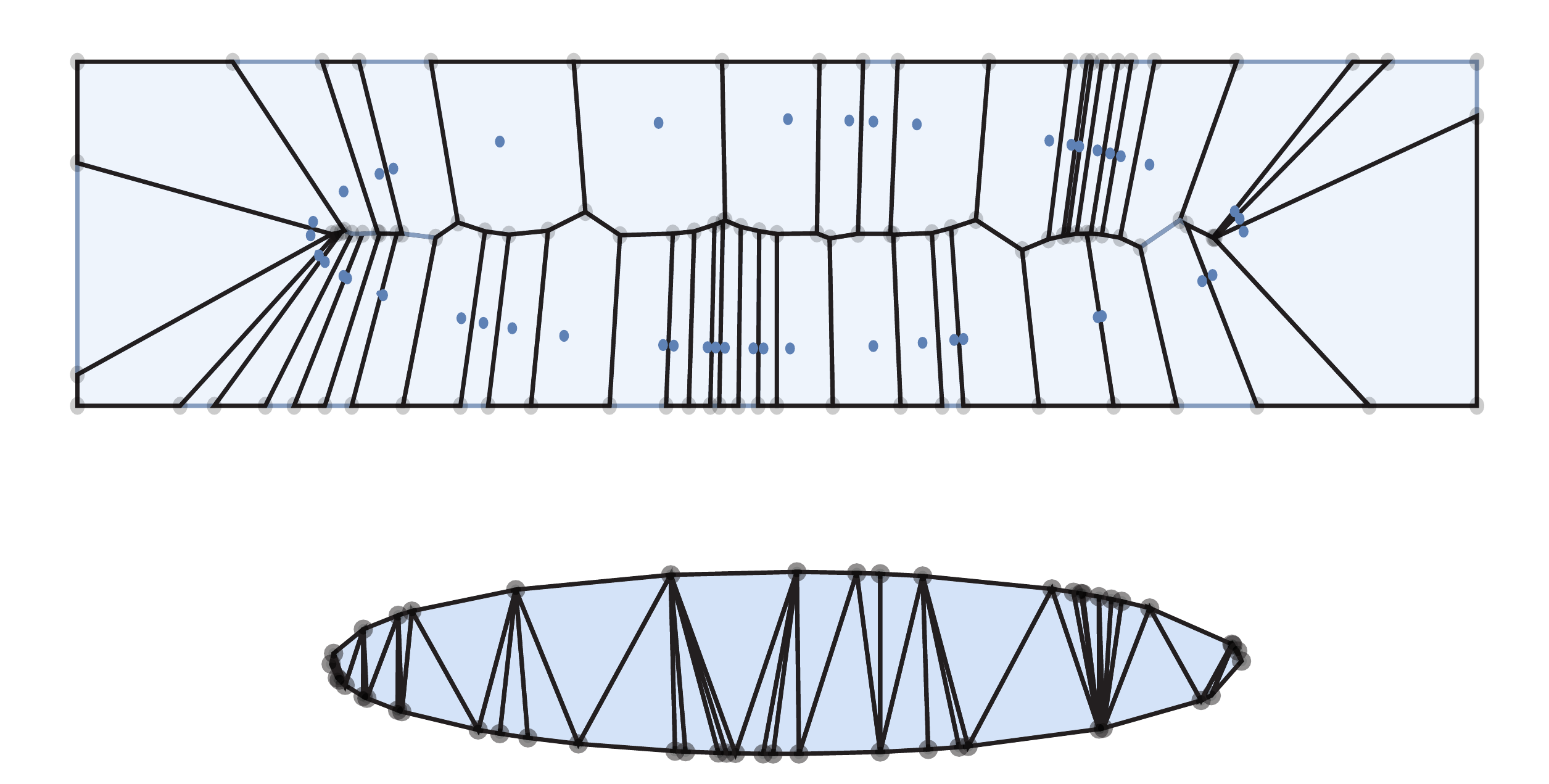}
\caption{Top: the Voronoi cells for some points sampled on a ellipse. Bottom: the corresponding Delaunay triangulation.}
\label{fig:delaunay_ellipse}
\vspace{-1em}
\end{figure}

It is interesting to compare Activated Simplices with methods which construct simplicial structures for manifold reconstruction, especially those that use the Delaunay complex \cite{Edelsbrunner:1983fe, Edelsbrunner:1998wx, Amenta:2000it, Edelsbrunner:2003wp, Dey:2003jf, Cheng:2005vy}. The Delaunay complex for a set of data points is a simplicial structure that is dual to the Veronoi cells for the points. For points in general position in the plane, three points are connected by a Delaunay triangle if their Voronoi regions meet at a point. Figure \ref{fig:voronoidelaunay} shows Voronoi boundaries and the corresponding Delaunay triangles for 6 points in the plane. Figure \ref{fig:delaunay_ellipse} shows the Voronoi cells and Delaunay triangles for 50 points sampled on a ellipse. The Delaunay complex in $d$ dimensions is defined analagously: $d+1$ points are connected by a $d$-dimensional simplex if their Voronoi regions intersect.

The Delaunay complex for points in $\R^d$ is a $d$ dimensional structure. It must be pared back to a lower dimensional structure to reconstruct a lower dimensional manifold. The methods that use Delaunay complexes differ in how this paring back is done. For example, $\alpha$-shapes \cite{Edelsbrunner:1983fe}  extracts sub-simplices of bounded diameter, whereas co-cone methods \cite{Amenta:2000it} select simplices by their proportions.

Activated Simplices doesn't construct a sub-complex of the Delaunay complex, but it is related to Delaunay complex methods in that it learns from the convex hull of the data in the sphere --- \cite{Edelsbrunner:1986wt} shows that the Delaunay complex can be constructed from the convex hull of the stereographic projection of data onto the sphere. The main difference between Activated Simplices and methods which pare back the  Delaunay complex is that the simplices are selected by Activated Simplices because they are activated by data, whereas in the Delaunay complex methods, sub-simplices of the complex are selected if they meet come geometric requirements such as size or proportion.

\section{Extremes and Archetypes}

\begin{figure}
\centering
\includegraphics[height=1.3in]{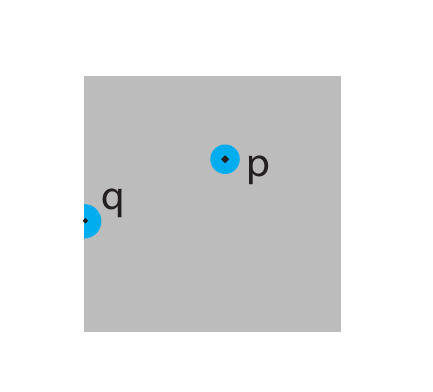}
\includegraphics[height=1.2in]{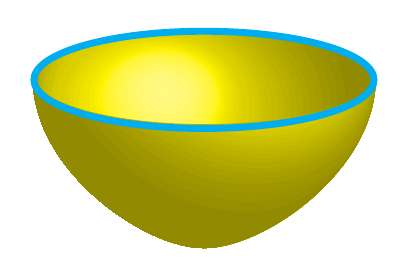}
\caption{ (a) Boundary points have a different neighborhood structure: $q$ is a boundary point on the closed square, but $p$ is not. (b) The blue circle is the boundary of the bowl.}
\label{fig:boundary_definition}
\vspace{-1em}
\end{figure}

\begin{figure}
\centering
\includegraphics[height=0.7in]{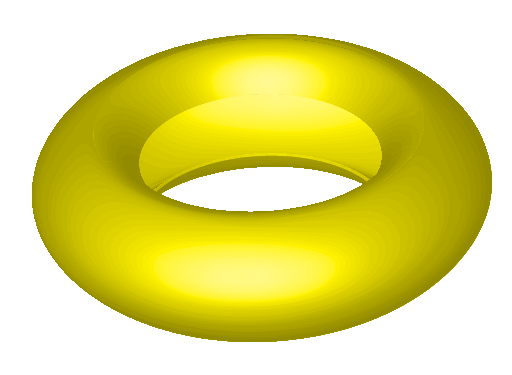}
\includegraphics[height=0.65in]{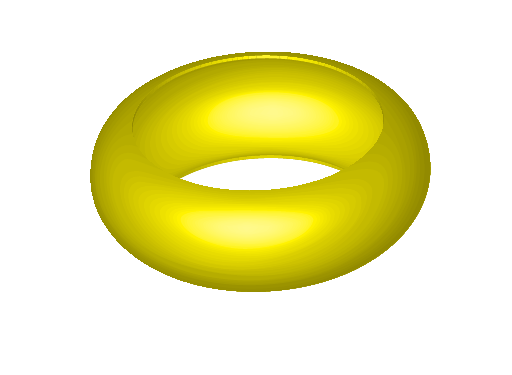}
\includegraphics[height=0.7in]{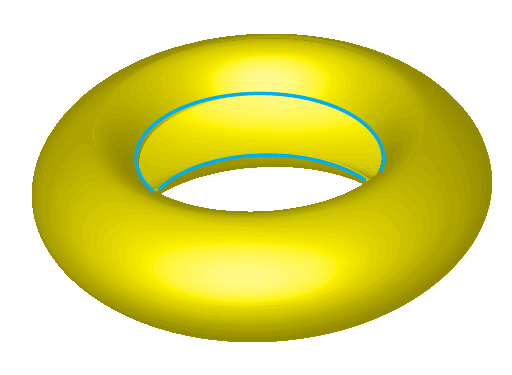}
\caption{Left: a car tire (not the inner tube, but the tire). Middle: the convex extremes of the manifold is the set of points where the tire might meet the road. Right: the two blue circles are the boundary points.}
\label{fig:boundary_tyre}
\vspace{-1em}
\end{figure}

We now discuss how extremes within data might be identified with Activated Simplices. But first we must define some notions of extreme.  A first notion is what might be called a \textit{convex extreme point}. This is a point on the data manifold that is not a convex combination of any other points on the manifold.  For example, in a cube, the corner points are convex extremes because they are not convex combinations of other points. On a 2-dimensional sphere, every point is a convex extreme point. This is an extrinsic notion of extreme --- a person looking on the manifold from afar can identify convex extremes, but a short sighted ant crawling on the manifold might notice nothing unusual as it crawls past a convex extreme. Figure \ref{fig:boundary_tyre} shows a tire and it's convex extremes. The convex extremes of a finite set of points are the vertices on the convex hull of the points.

Archetypal Analysis \cite{cutler1994archetypal} \cite{chen2014fast} learns a convex model for the data (it learns bases such that the convex hull encloses the data). The bases that are learned are approximations to extreme points.

A second notion of extreme point comes from the mathematical notion of a manifold with boundary. In this formulation an interior point has neighborhoods homeomorphic to disks, but boundary points have neighborhoods homeomorphic to half disks. Figure \ref{fig:boundary_definition} (a) shows a neighborhood for an interior point $p$ and a boundary point $q$ on the square. Figure  \ref{fig:boundary_definition} (b) shows the boundary points of the bowl and Figure \ref{fig:boundary_tyre} shows the boundary points of the tire.

Boundary point is an intrinsic notion --- a short sighted ant will notice that the neighborhood is different at a boundary point. Boundary points correspond to extreme configurations. For example, a pose, where a limit of the range of motion at some joint is reached, is a boundary point on the manifold of poses.

Figure \ref{fig:boundary_tyre} illustrates that convex extremes and boundary points may not coincide.

\subsection{Boundaries and Activated Simplices}

\begin{figure}
\centering
\includegraphics[height=1.2in]{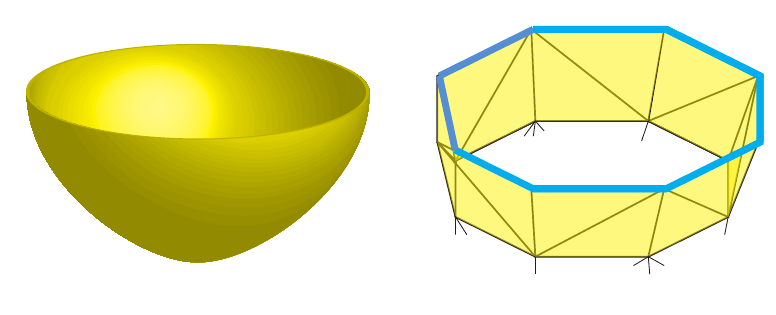}
\caption{The bowl and the top part of a triangulation, with boundary simplices in blue.}
\label{fig:bow_triangulation}
\vspace{-1em}
\end{figure}

\begin{figure}
\centering
\includegraphics[height=1.2in]{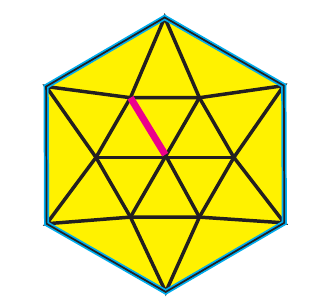}
\caption{The red line is an interior simplex, the blue segments are boundary simplices}
\label{fig:boundary_simplices_3}
\vspace{-1em}
\end{figure}

We now develop a notion of boundary for a simplicial structure.
Figure \ref{fig:bow_triangulation} shows the bowl and part of a triangulation. The blue segments are boundary simplices. Figure \ref{fig:boundary_simplices_3} shows another simplicial structure; the blue simplices are boundary simplices, the red segment is an internal segment. These pictures suggest a definition of boundary simplex --- a simplex is a boundary simplex if it has co-dimension 1 in a maximal simplex and it is not a sub-simplex of two maximal simplices. In Figure \ref{fig:boundary_simplices_3} the blue segments are sub-simplices of only one maximal simplex, whereas the red segment is a sub-simplex of two maximal simplices.

\begin{figure}
\centering
\includegraphics[height=3.5in]{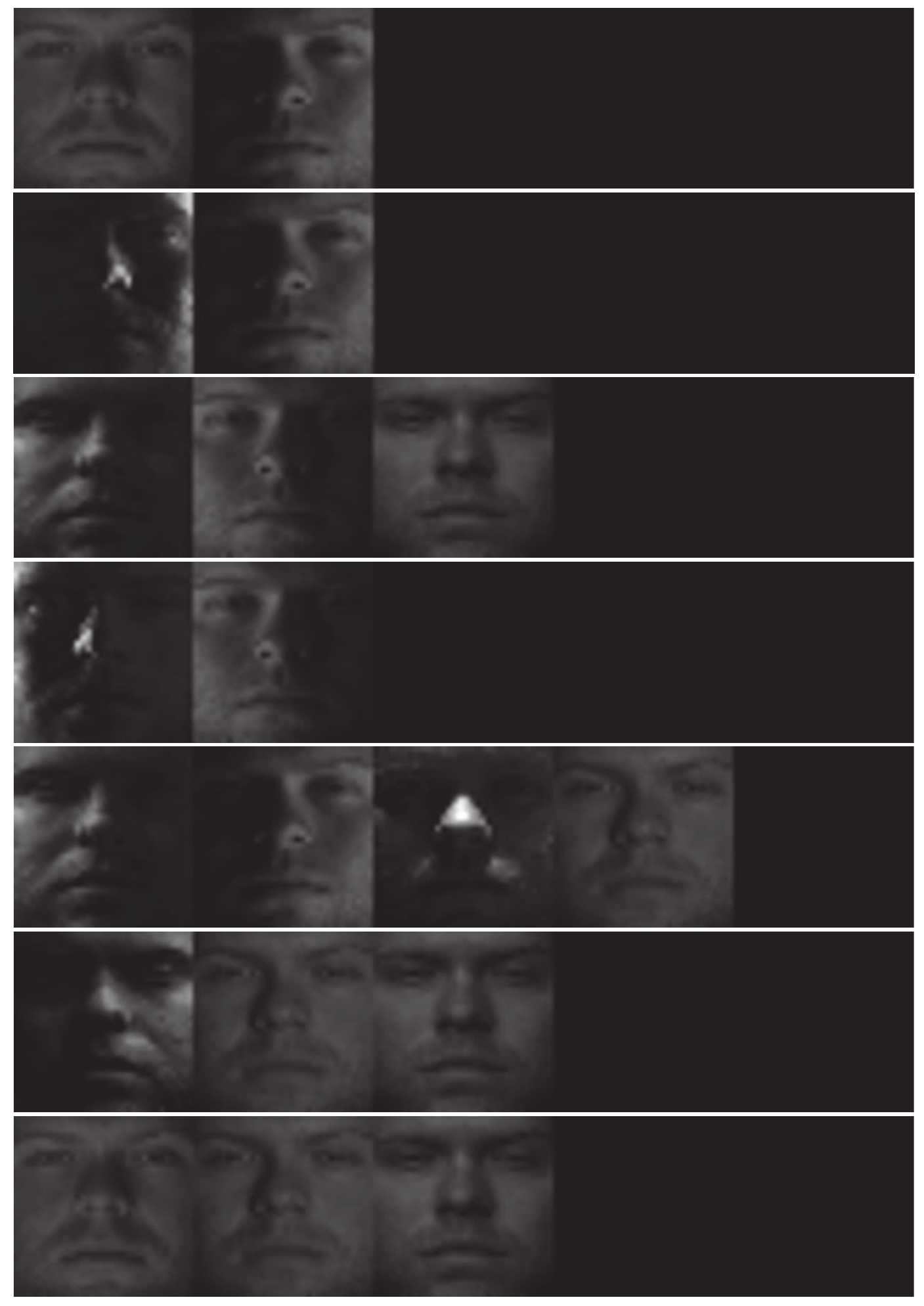}
\caption{Each show shows a boundary simplex found by our method. We can see that, for most case, there is at least an ``extreme'' lighting condition in the simplex.}
\label{fig:boundary_face}
\vspace{-1em}
\end{figure}

We can use this notion to detect extremes using Activated Simplices. Figure \ref{fig:boundary_face} shows some boundary faces found in the Yale B face data. We used  $64$ face images of the same person under different lighting conditions. We learn $10$ bases, obtain $27$ simplices and identify $14$ maximal simplices. Seven of the maximal simplices are in the interior of the manifold. It means all of their co-dimension 1 sub-simplices are sub-simplices of other maximal simplices. Each of the remaining seven maximal simplices has one co-dimension 1 sub-simplex that is a boundary.  Figure \ref{fig:boundary_face} shows vertices for the seven boundary simplices. We can see that the boundary simplices include the images corresponding to extreme lighting directions.

The utility of the boundaries found from Activated Simplices depends on the data manifold and on the simplicial structure that is constructed. If the simplicial structure is stringy and lumpy there will be many boundary simplices, and it will be hard to make sense of these extremes. This kind of simplicial structure can arise because it reflects the reality of the data manifold, or it can occur when the data manifold is under-sampled.

\subsection{Comparison with Archetypal Analysis}
\begin{figure}
\centering
\includegraphics[height=1in]{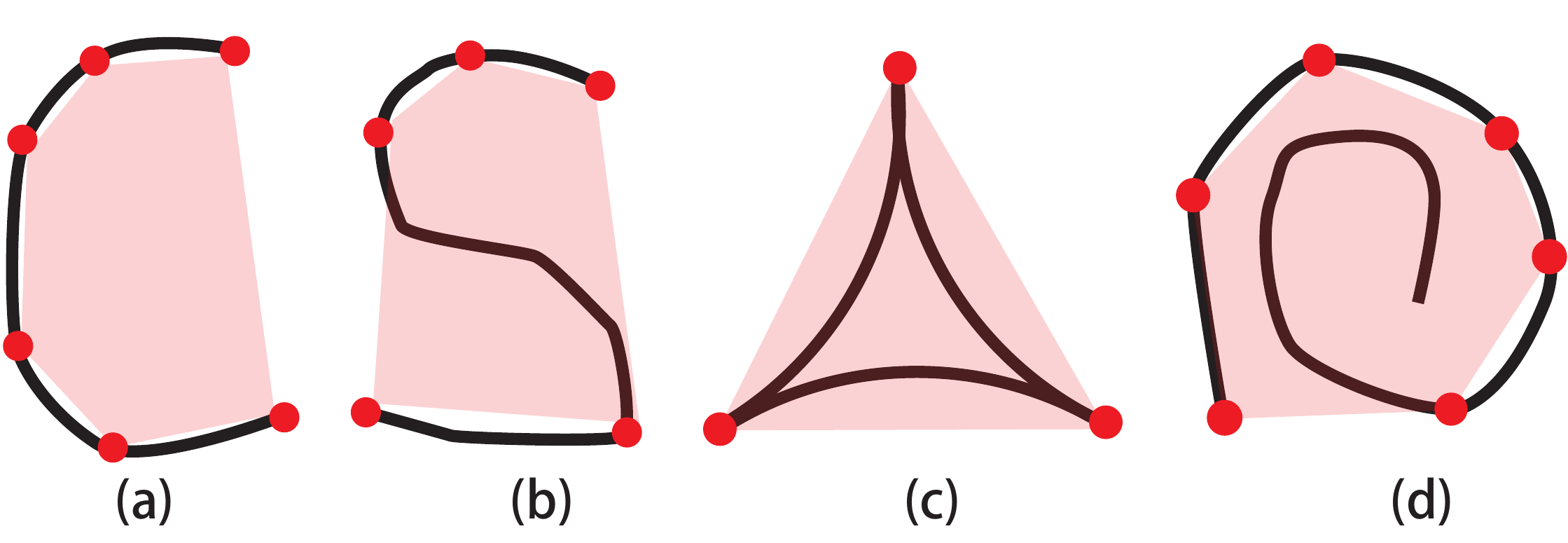}
\caption{Bases that might be learned with Archetypal Analysis. The black line represents the data manifold and the red dots are the learned bases. (a) Archetype analysis will obtain similar bases as our approach when data is on boundary of convex set. (b,c,d) Archetype Analysis doesn't model details away from convex extremes. Note also that the regions that are represented by the bases (the convex hull of the bases) are much larger than the manifolds from which they are learned. }
\label{fig:archetype}
\vspace{-1em}
\end{figure}

Archetypal Analysis learns a global convex model for the data. This is appropriate when the data manifold is a convex subset of Euclidean space. In other situations the Archetypal Analysis fails to learn the details of the data. Figure \ref{fig:archetype} shows some structures that might be constructed by Archetypal Analysis.

\begin{figure}
\centering
\includegraphics[height=1in]{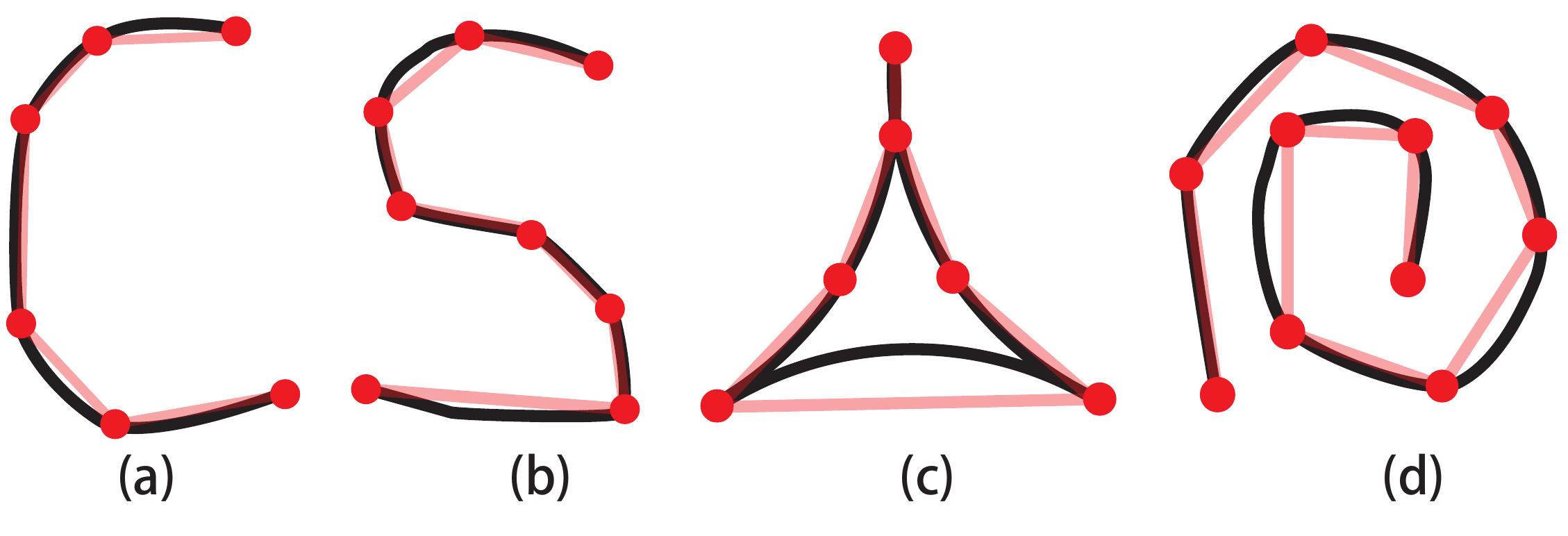}
\caption{Bases that might be learned by activated simplices. The black line represents the data manifold and the red dots are the learned bases. The red segments are the activated simplices.}
\label{fig:simplices}
\vspace{-1em}
\end{figure}

\begin{figure*}
\includegraphics[height=7in]{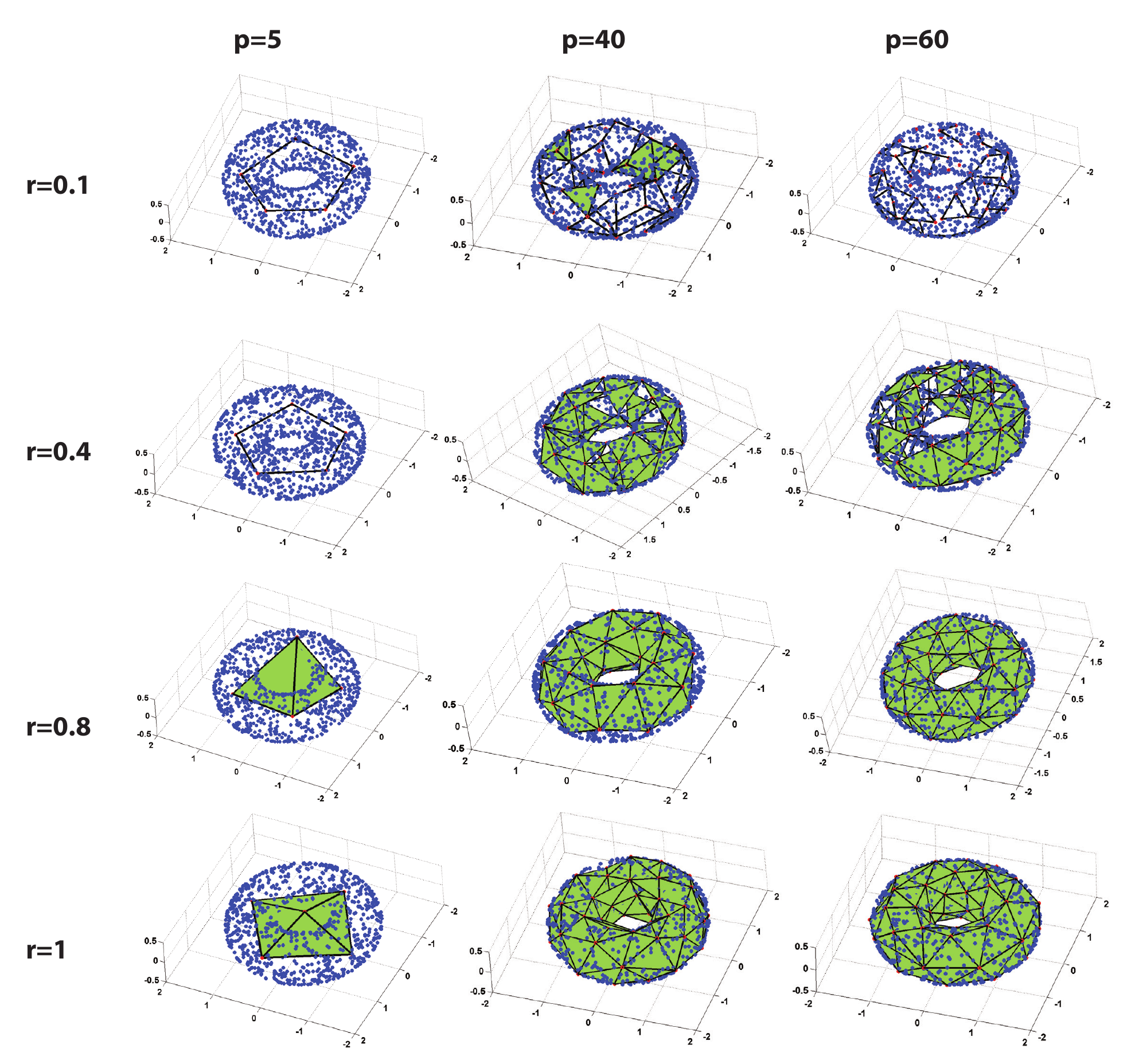}
\caption{Simplicial structures learned from data sampled from the torus, using $p = 4, 40$ and $60$ bases and using $r = 0.1, 0.4, 0.8$ and $1$. $r=0.1$ is heavily penalized and $r=1$ is unpenalized.}
\label{fig:torus_reconstructions}
\end{figure*}

In contrast, Activated Simplices learns a local convex model for the data.  Figure \ref{fig:simplices} shows structures that might be learned by Activated Simplices.

Archetypal Analysis learns convex extremes but it can fail to find intrinsic boundaries. Activated Simplices can find convex extremes among the bases (though not all bases will be extremes), and it can find intrinsic boundaries when the manifold is well reconstructed.

{\small
\bibliographystyle{ieee}
\bibliography{egbib}
}

\end{document}